\documentclass[sigconf]{acmart}

\usepackage{amsmath,amsfonts,amsthm} 
\usepackage{bbm}
\usepackage{textcomp}
\usepackage{xcolor}
\usepackage{algorithm, algorithmic}
\usepackage{multirow}
\usepackage{enumitem} 
\usepackage{bm}
\usepackage{booktabs}

\def\done{\hspace*{\fill} \rule{1.8mm}{2.5mm} \\}
\newtheorem{assumption}{\hspace{-0.16 in} {\bf Assumption} }
\newtheorem{proposition}{\hspace{-0.16 in} {\bf Proposition} }
\newtheorem{theorem}{\hspace{-0.16 in} {\bf Theorem} }
\newtheorem{lemma}{\hspace{-0.16 in} {\bf Lemma}}
\newtheorem{definition}{\hspace{-0.16 in} {\bf Definition}}

\AtBeginDocument{%
  }

\setcopyright{acmcopyright}
\copyrightyear{2023} 
\acmYear{2023} 
\setcopyright{acmlicensed}\acmConference[WWW '23]{Proceedings of the ACM Web Conference 2023}{April 30-May 4, 2023}{Austin, TX, USA}
\acmBooktitle{Proceedings of the ACM Web Conference 2023 (WWW '23), April 30-May 4, 2023, Austin, TX, USA}
\acmPrice{15.00}
\acmDOI{10.1145/3543507.3583456}
\acmISBN{978-1-4503-9416-1/23/04}

\begin{document}

\title{Interactive Log Parsing via Light-weight User Feedback}
\author{Liming Wang}
\affiliation{%
   \institution{Chongqing University}
   \city{}
   \state{}
   \country{}}
\email{wlm\_1203@163.com}

\author{Hong Xie}
\affiliation{%
   \institution{Chongqing University}
   \city{}
   \state{}
   \country{}}
\email{xiehong2018@foxmail.com}

\author{Ye Li}
\affiliation{%
   \institution{Alibaba}
   \city{}
   \state{}
   \country{}}
\email{liye.li@alibaba-inc.com}

\author{Jian Tan}
\affiliation{%
   \institution{Alibaba}
   \city{}
   \state{}
   \country{}}
\email{j.tan@alibaba-inc.com}

\author{John C.S. Lui}
\affiliation{%
   \institution{The Chinese University of Hong Kong}
   \city{}
   \state{}
   \country{}}
\email{cslui@cse.cuhk.edu.hk}
\renewcommand{\shortauthors}{Wang et al.}

\begin{abstract}
Template mining is one of the foundational tasks to support log analysis, 
which supports the diagnosis and troubleshooting of large scale Web applications.  
This paper develops a human-in-the-loop template mining framework to support 
interactive log analysis, which is highly desirable in 
real-world diagnosis or troubleshooting of Web applications 
but yet previous template mining algorithms fail to support it.  
We formulate three types of light-weight user feedback 
and based on them we design three atomic human-in-the-loop template mining algorithms.  
We derive mild conditions under which the outputs of 
our proposed algorithms are provably correct.  
We also derive upper bounds on the computational complexity and 
query complexity of each algorithm.  
We demonstrate the versatility of our proposed algorithms by combining 
them to improve the template mining accuracy of five representative algorithms over 
sixteen widely used benchmark datasets.  
\end{abstract}



\maketitle


\section{Introduction}
 
With the growing scale and complexity of Web applications such as 
cloud computing and micro-service systems 
\cite{meng2020semantic,huo2021semparser,liu2022uniparser}, 
system event logs (we call them logs for brevity) provide first-hand information for engineers to monitor the health status 
of the system and troubleshoot \cite{he2021survey}.  
The raw logs are of a vast volume containing much redundant information, 
making it difficult for engineers to analyze them.  
Template mining is one of the foundational tasks to support log analysis.  
It aims to partition logs into clusters such 
that similar logs are in the same cluster \cite{he2021survey}.  
It also extracts a ``template'' for each cluster, 
which summarizes the key information of  the logs in a cluster \cite{he2021survey}.  
   
A number of template mining algorithms were proposed  \cite{boyagane2022vue4logs,sedki2021awsom,plaisted2022dip,rucker2022flexparser,liu2022uniparser,he2021survey}, 
which enable the automatic extraction of templates.  
However, previous template mining algorithms do not support 
interactive log analysis, which is highly desirable in 
real-world diagnosis or troubleshooting of Web applications.  
In particular, in the diagnosis or troubleshooting, 
engineers may have varied granularity on the clustering or semantics of templates.  
As she/he dives deeper into the diagnosis or troubleshooting, 
higher clustering or semantic granularity on susceptible logs 
may be preferred, 
while lower clustering or semantic granularity on irrelevant logs is preferred.  
Higher clustering or semantic granularity can be achieved by splitting a cluster 
into several smaller clusters with larger inner similarities 
and extracting templates with richer semantics accordingly, 
while lower clustering granularity can be achieved by 
merging several similar clusters and 
extracting templates with less semantics accordingly.  

We design a human-in-the-loop template mining framework to enable 
interactive log analysis.  
We do not extract templates from scratch; 
instead, we solicit user feedback to adjust 
the templates extracted by a base algorithm 
toward the user's needs or preferences.  
Our framework is generic to be deployed on the output of any previous template mining algorithms. 
Our framework supports three atomic human-in-the-loop operations: 
(1) improving the richness of the semantics of a template; 
(2) merging merge two clusters;
(3) splitting split a cluster.   
To relieve the user's burden in providing feedback, 
we consider three types of light-weight feedback.    
The first one is indicating whether a given template has all semantics that the user itself wants.  
The second one is selecting tokens from a given template that the 
engineer does not care, 
which we call dummy tokens.  
The third one is selecting a template from a given set that has the same semantics as the given template.  
We design computationally efficient algorithms that 
creatively combine these three types of feedback 
to implement three desired  atomic human-in-the-loop operations.  
Our algorithms work in a stream fashion in the sense that it only needs 
to pass the input log data once.  
Furthermore, we prove upper bounds on the computational complexity and 
query (seek user feedback) complexity, 
which reveal a fundamental understanding of the efficiency of our proposed algorithms.   
Finally, we demonstrate the application of algorithms by applying 
them to improve the template mining accuracy of five representative algorithms over 
sixteen widely used benchmark datasets.  
The highlights of our contributions include: 
\begin{itemize}
\item 
Formulation of three types of light-weight user feedback 
and three atomic human-in-the-loop operations.  

\item 
Upper bounds on the computational complexity and 
query (seek user feedback) complexity of proposed algorithms.  

\item 
Extensive experiment evaluation on sixteen benchmark data-\\sets.  
\end{itemize}

 
\section{Releated Work}

Previous works on log parsing can be categorized into two lines:  
(1) pattern aware log parsing, which extracts frequent patterns of logs; 
(2) semantic aware log parsing, which extracts templates containing key semantics of logs.

\subsection{Pattern Aware Log Parsing}

Clustering-based log parsing methods follow the workflow of 
clustering logs and then extracting templates of each cluster.  
LKE \cite{fu2009execution} extracts raw keys of logs by applying empirical rules to erase the parameters of logs.  
The similarity between logs is quantified by the edit distance between their raw keys.  
Based on this similarity measure, logs are clustered into different groups, 
and the common part of raw keys serves as the template of a group.  
IPLoM \cite{makanju2009clustering} utilizes hierarchical clustering to partition logs 
and then produce the descriptions of each cluster, i.e., the template for each cluster.   
LFA \cite{nagappan2010abstracting}  applies the Simple
Log file Clustering Tool for log abstraction.   
LogMine \cite{hamooni2016logmine} 
is composed of a clustering module and a pattern recognition module. 
Its novelty lies in the efficient computational implementation of these two modules in the map-reduce framework.   
CLF \cite{zhang2019efficient} extracts templates via heuristic rules, 
i.e., clustering logs based on heuristic rules, adjusting the clustering based on separation rules, 
and finally generating a template for each cluster.  
Inspired by word2vec, LPV \cite{xiao2020lpv} uses deep learning to 
vectorize logs, cluster logs based on vector similarity, 
and extract templates from the resulting clusters.  
Vue4logs \cite{boyagane2022vue4logs}
uses a vector space model to extract event templates, which vectorize log and group logs based on their vector similarity.  
Character and length-based filters are used to extract templates.  

Frequency-based methods rely on intuition that frequent patterns are more likely to be templates.  
Ft-tree \cite{zhang2017syslog} 
identify frequent combinations of (syslog) words as templates of logs.  
It supports incremental learning of log templates.  
Logram \cite{dai2020logram} 
utilizes the frequency of n-gram dictionaries to parse logs, 
where frequent n-gram dictionaries are identified as templates.   
It supports online parsing of logs.  
Meting \cite{coustie2020meting} is a parametric log parser, 
which is also built on frequent n-gram mining.  
AWSOM-LP \cite{sedki2021awsom} organizes logs into patterns via a simple text processing method.
It then applies frequency analysis to logs of the same group to identify static and dynamic content of log events.

Tree-based methods design different trees to encode different log parsing rules.  
Drain \cite{he2017drain} uses a fixed depth parse tree to extract templates of logs.   
This fixed depth parse tree encodes specially designed parsing rules. 
Prefix Graph \cite{chu2021prefix} 
is a probabilistic graph structure, which is an extension of a prefix tree. 
Two branches are merged together when they have high similarity in the probability distribution.   
The combination of cut-edges in root-to-leaf paths of the graph.  
USTEP \cite{vervaet2021ustep} 
uses an evolving tree structure to extract the template of logs.  
It is an online log parsing method.  
DIP 2022 \cite{plaisted2022dip} 
is a tree-based log parser. 
The primary methodological innovation is that 
DIP considers the actual tokens at which the two messages disagree and 
the percentage of matching tokens.   

A number of works applied deep learning to parse logs. 
DeepLog \cite{du2017deeplog} 
treats a log as a natural language sequence and
applies Long Short-Term Memory (LSTM) to extract templates.  
LogPhrase \cite{meng2020logparse} 
casts the template extraction problem as a word classification problem.  
It applies deep learning to learn the features of static words and variable words. 
Nulog \cite{nedelkoski2020self} 
casts the parsing task as a masked language modeling (MLM) problem 
and uses a self-supervised learning model to address it.  
UniLog 2021 \cite{zhu2021unilog}
casts the log analysis problem as a multi-task learning problem and 
proposes a log data pre-trained transformer to parse logs.
LogDTL \cite{nguyen2021logdtl} is a semi-supervised method.  
It uses a transfer learning technique together with the deep neural network to balance the trade-off between 
the accuracy of the extracted template and human resources for manual labeling. 
FlexParser \cite{rucker2022flexparser} 
trains a stateful LSTM to parse logs. 

We are also aware of the following notable methods that do 
not belong to the above types.  
Spell \cite{du2016spell} 
is an online streaming template mining algorithm.  
It extracts templates of logs via a longest common subsequence-based approach. 
Logan \cite{agrawal2019logan}  
is a distributed online log parser, which is also based on the
Longest Common Subsequence algorithm.  
LogPunk 2021 \cite{zhang2021efficient} 
and QuickLogS \cite{fang2021quicklogs} are two notable hash-like methods for log parsing.   
LogStamp \cite{tao2022logstamp}
is a sequence labeling-based automatic online log parsing method.     
MoLFI \cite{messaoudi2018search} casts the log message identification problem as a multi-objective problem.  
It applies evolutionary approaches to solve this problem.  
Paddy \cite{huang2020paddy}  uses a dynamic dictionary structure to build an inverted index, 
which enables an efficient search of the template candidates.  
AECID-PG \cite{wurzenberger2019aecid} 
is a density-based log parsing method.

\subsection{Semantic Aware Log Parsing}

Recently, a few works brings attention to the semantics of templates 
\cite{meng2020semantic,huo2021semparser,liu2022uniparser}.  
These methods apply deep learning to enrich the semantics of templates, 
which require a large amount of training data.  
Unlike these works, we utilize light-weight human feedback 
to adjust the log mining results.  
Through this, we not only enrich the semantics of templates 
but also improve the accuracy of the group of logs.  
Furthermore, we have a rigorous analysis of the correctness, 
computational complexity, and human feedback query complexity.  
This aspect is missed in most previous works.


\section{\bf Model and Problem Formulation}

\subsection{\bf Characterizing Logs and Templates}

To simplify notations, for each integer $I \in \mathbb{N}_+$, 
we define $[I]$ as 
$
[I] \triangleq \{1, \ldots, I \}.  
$
Let $\mathcal{D}$ denote a dictionary of tokens.  
We characterize each log as a sequence of tokens.  
To facilitate the presentation, 
we define the following operations regarding sequences of tokens.     

\begin{definition}
	Given two sequences $\bm{a} {=} (\! a_1, \ldots, a_I \!)$ 
	and $\bm{b} {=} (\! b_1, \ldots, b_J \!)$, 
	where $a_i, b_j \in  \mathcal{D}, \forall i \in [I], j \in [J]$:  
	\begin{itemize}
		\item 
		Equal `$=$': 
		$\bm{a}$ and $\bm{b}$ are equal denoted by 
		$
		\bm{a} = \bm{b}, 
		$
		if and only if $I=J$ and $a_i = b_i, \forall i \in [I]$.  
		
		\item 
		Subsequence `$\sqsubseteq$': 
		$\bm{a}$ is a subsequence of $\bm{b}$ denoted by 
		$
		\bm{a} \sqsubseteq \bm{b}, 
		$
		if and only if 
		there exits a sequence $1 \leq j_1 < j_2 < \ldots < j_I < J$ 
		such that $a_i = b_{j_i}, \forall i \in [I]$.  
		\item $\texttt{len}(\cdot)$:  $\texttt{len}(\bm{a}) = I$.   
		\item $\texttt{LCS}(\cdot, \cdot)$: $\texttt{LCS}(\bm{a}, \bm{b}) = $ the longest common subsequence between $\bm{a}$ and $\bm{b}$.  
	\end{itemize}
\end{definition}
To facilitate labeling, 
we use the $<$*$>$ symbol to replace all tokens except that in the templates and allow one $<$*$>$ to represent several tokens.  
It does not affect the original semantic representation of the template.  
We consider a set of $N \in \mathbb{N}_+$ logs to be parsed.  
Let $\bm{L}_n$ denote log $n \in [N]$, which is a sequence of tokens from 
the dictionary $\mathcal{D}$, formally 
$
\bm{L}_n
\triangleq  
(L_{n,1}, \ldots, L_{n,M_n}),  
$
where $M_n \in \mathbb{N}_+$ denote the length of log $n$ 
and $L_{n,m}  \in  \mathcal{D}, \forall m \in [M_n]$.   
All these $N$ logs are partitioned into $K \in \mathbb{N}_+$ disjoint clusters based on their message or semantics.  
Let $\mathcal{C}_k \subseteq [N]$, where $k \in [K]$ denote the set of indexes of 
the logs in cluster $k$.  
These $K$ clusters are full,  i.e., 
$
\cup_{k  \in [K]}  \mathcal{C}_k = [N].  
$
This property captures each log that belongs to at least one cluster.  
Furthermore, these $K$ clusters are disjoint, i.e., 
$
\mathcal{C}_k \cap \mathcal{C}_{k'} = \emptyset, 
\forall k, k' \in [K] \text{ and } k \neq k'.  
$
This property captures that there is no log that belongs to more than one cluster.  
In other words, there is no message redundancy or message 
ambiguity in the clusters.  
Cluster $k$, where $k \in [K]$, is associated with a template $\bm{T}_k$, 
which captures the message or semantics of cluster $k$.  
The template $\bm{T}_k$ is a common subsequence of the logs that 
belong to cluster $k$, i.e., 
$
\bm{T}_k 
\sqsubseteq 
\bm{L}_n, 
\forall n \in \mathcal{C}_k.  
$
Note that the template $\bm{T}_k$ is not necessarily the longest common subsequence.   
For example, consider a log cluster with two logs "Failed password from port 11, user=root" and "Failed password from port 12, user=root". The template of this cluster is "Failed password from port $<$*$>$ user $<$*$>$" but not "Failed password from port $<$*$>$ user root".

The clusters $\mathcal{C}_k, \forall k \in [K]$ and templates $\bm{T}_k, \forall k \in [K]$ are 
essential for supporting downstream applications such as anomaly detection, 
root cause analysis, etc.  
We impose the following natural assumption on templates 
to simplify the discussion.  
\begin{assumption}
	There does not exist two templates $\bm{T}_k$ and $\bm{T}_{k'}$, 
	where $k, k' \in [K]$ and $k \neq k'$, 
	such that $\bm{T}_k \sqsubseteq \bm{T}_{k'}$.   
	\label{ass:model:templateSubset}
\end{assumption}

\noindent 
Assumption \ref{ass:model:templateSubset} captures that 
there are no templates whose message is part of another template.  
It ensures that each template contains a new message compared with 
the other.  

\noindent 
{\bf Remark.}  
The clusters $\mathcal{C}_k, \forall k \in [K]$ and templates $\bm{T}_k, \forall k \in [K]$  
are the ground truth, 
and they are defined by user preference or needs.    
This ground truth may vary across different users, 
as different users may have different preferences over the 
semantics of templates.   
For example, different users may prefer different granularity on the templates.  
Even for the same user, she/he may prefer a low granularity when the system 
is at normal status while preferring a high granularity in abnormal status.  

\subsection{\bf Characterizing A Log Mining Algorithm}

A number of rule-based or machine learning-based algorithms 
aim to recover the clusters and templates automatically.  
We present a unified way to characterize them through their output.  
Formally, let $\widehat{\mathcal{C}}_1, \ldots, \widehat{\mathcal{C}}_{\widehat{K}}$ 
denote the clusters extracted by a log mining algorithm, 
where $\widehat{K} \in \mathbb{N}_+$, 
which satisfy 
\[
\cup_{k  \in [\widehat{K}]}  \widehat{\mathcal{C}}_k = [N], \quad
\widehat{\mathcal{C}}_k \cap \widehat{\mathcal{C}}_{k'} = \emptyset,  
\forall k, k' \in [K] \text{ and } k \neq k'. 
\]
Namely, the mined clusters satisfy the full property and disjoint property.  
Note that $\widehat{K}$ can be greater, equal, or smaller than $K$, 
depending on the selected algorithm and hyperparameter selection.  
Let $\widehat{\bm{T}}_k$ denote the template associated with 
cluster $\widehat{\mathcal{C}}_k$, where $k \in [\widehat{K}]$.  
We call $\widehat{\mathcal{C}}_k$ and $\widehat{\bm{T}}_k$,  
where $\forall k\in[\widehat{K}]$ is the 
mined cluster and mined template.  
The mined templates satisfy Assumption \ref{ass:model:templateSubset}.  

The following notion characterizes the errors 
at the cluster level.   

\begin{definition} 
	A mined cluster $\widehat{\mathcal{C}}_k$, where $k \in [\widehat{K}]$, 
	is pure, if there exists $k' \in [K]$ such that 
	\begin{equation}
		\widehat{\mathcal{C}}_k 
		\subseteq 
		\mathcal{C}_{k'},  
		\label{eq:model:pure}
	\end{equation}
	otherwise it is mixed.  
	A pure cluster is full, if the equality in (\ref{eq:model:pure}) holds, 
	otherwise it is partial.  
	\label{def:model:pure}
\end{definition}  

\noindent
Definition \ref{def:model:pure} states that a pure mined cluster 
contains only one type (the type is defined concerning the ground truth template 
associated with it) of logs. 
A mixed mined cluster contains more than one type of logs.  
A mixed mined cluster indicates an error in the cluster level.  
A partial pure cluster also indicates an error in the cluster level.
For example, consider three mined clusters
	$\widehat{\mathcal{C}}_1=\{\bm{L}_1\}$, 
	$\widehat{\mathcal{C}}_2=\{\bm{L}_2\}$, 
	$\widehat{\mathcal{C}}_3=\{\bm{L}_3, \bm{L}_4\}$, and ground truth clusters ${\mathcal{C}}_1=\{\bm{L}_1,\bm{L}_2\}$, 
	${\mathcal{C}}_1=\{\bm{L}_3\}$, 
	${\mathcal{C}}_1=\{\bm{L}_4\}$. 
	Under this definition, all mined clusters are inaccurate. 
	And $\widehat{\mathcal{C}}_3$ is a mixed mined error, while $\widehat{\mathcal{C}}_1$  and $\widehat{\mathcal{C}}_1$  are partial pure errors.  
As with \cite{zhu2019tools}, we use the group accuracy (GA) metric to quantify the clustering accuracy, formally: 
\[
\mathrm{GA} 
\triangleq 
\frac{ 1 }{ N }
\sum\nolimits_{ k \in [\widehat{K}] }  
\bm{1}_{ \{\exists k' \in [K], \mathcal{C}_{k'} = \widehat{C}_k \} }  
|   \widehat{C}_k |
\]
The following definition characterizes
message level errors.  
\begin{definition} 
	A mined template $\widehat{\bm{T}}_k$, where $k \in [\widehat{K}]$, 
	has complete message, if there exists $k' \in [K]$ such that 
	$
	\bm{T}_{k'} 
	\sqsubseteq
	\widehat{\bm{T}}_k.  
	$ 
	Otherwise it has message loss.   
	\label{def:model:message}
\end{definition}

\noindent
Definition \ref{def:model:message} states that a mined template 
has the complete message if it contains a ground truth template 
as a subsequence.  
Namely, at the message level, it has the full message of 
a ground truth temple.  
Otherwise, it has message loss.  
In other words, it does not contain the full message 
of any ground truth template.  
For example, consider a ground truth template $\bm{T}_1=$
	"Failed password from port $<$*$>$ user $<$*$>$". 
	Two mined templates are $\widehat{\bm{T}}_1=$
	"Failed password from port $<$*$>$" and 
	$\widehat{\bm{T}}_2=$
	"Failed password from port $<$*$>$ user root". 
	Even though both mined templates are inconsistent with the ground truth template, $\widehat{\bm{T}}_1$ loses critical information about the user, while $\widehat{\bm{T}}_2$ only has partial data redundancy and no semantic information loss.  
A template with message loss may not support 
the downstream applications well.  
Meanwhile, if a mined cluster is pure, although a mined template with the complete message may contain some redundancy, 
this redundancy does not distract the user a lot.  
Thus we focus on templates with message loss.  
Based on this, we propose the message accuracy (MA) metric to quantify the message level accuracy of templates, formally: 
\[
\mathrm{MA} 
{\triangleq} 
\!\frac{1}{N} 
\sum\nolimits_{ k \in [\widehat{K}] } 
\sum\nolimits_{ n \in \widehat{C}_k }
\bm{1}_{\{
\text{ground-truth template of $\bm{L}_n$} 
\sqsubseteq
\widehat{T}_k
\} 
}
\]

The following proposition states that 
a mixed-mined cluster has message loss 
in its associated mined template.   

\begin{proposition}
	If a cluster is mixed, then the template associated with it 
	has message loss.  
\end{proposition}

\subsection{\bf Problem Formulation}
 
Connecting mined clusters with templates, 
the notions defined in Definitions \ref{def:model:pure} 
and \ref{def:model:message} enable us to classify the 
errors into the following three types:   
(1) {\bf Loss-pure error}, which corresponds to that 
a cluster is pure but its associated template has message loss.  
(2) {\bf Complete-partial error}, which corresponds to that a cluster is partial, 
but the associated template has the complete message.  
(3) {\bf Loss-mixed error}, 
which corresponds to a mixed cluster and its associated template has message loss.  
\textit{
Our objective is to design a human-in-the-loop algorithm 
to eliminate these three types of errors. 
}


\section{Eliminating Loss-pure Error}


\subsection{Human Feedback Model}

We consider three types of lightweight user feedback, 
which are provided based on users' perception of the message of token sequences.    
Algorithm \ref{Algo:HumanFeedback}  summarizes the procedures that 
we design to solicit such user feedback.

\begin{algorithm}[htb]
\caption{\texttt{Human Feedback} } 
\label{Algo:HumanFeedback} 
\begin{algorithmic}[1]
\STATE
\textbf{SubFunction} \texttt{Human-Message-Loss} ($\widehat{\bm{T}}$)
\label{Algo:Human-message}
\STATE
\ \ \ \ \ \
Present the template $\widehat{\bm{T}}$ to the user
\STATE
\ \ \ \ \ \
\textbf{return} user feedback 1 (message loss) or 0 (no loss)
\STATE
\textbf{SubFunction} \texttt{Human-Dummy-Token} ($\widehat{\bm{T}}$)
\STATE
\ \ \ \ \ \
Present the template $\widehat{\bm{T}}$ to the user
\STATE
\ \ \ \ \ \
The user selects at least one dummy token
\STATE
\ \ \ \ \ \
\textbf{return} the selected dummy tokens
\STATE
\textbf{SubFunction} \texttt{Human-Select} ($\widehat{\bm{T}}, \mathcal{T}$)
\STATE
\ \ \ \ \ \
For each template in $\mathcal{T} $, extract the LCS between $\widehat{\bm{T}}$ and it
\label{Step:HumanSelect:LCS}
\STATE
\ \ \ \ \ \
Sort templates in $\mathcal{T} $ based on the length of extracted LCS
\begin{flushleft}
\ \ \ \ \ \
in descending order
\end{flushleft}
\label{Step:HumanSelect:Sort}
\STATE
\ \ \ \ \ \
Delete all templates with zero length extracted LCS and
\begin{flushleft}
\ \ \ \ \ \
present the remaining sorting list to the user
\end{flushleft}
\label{Step:HumanSelect:Delete}
\STATE
\ \ \ \ \ \
\textbf{return} the selected template or null (if none is selected)
\end{algorithmic}
\end{algorithm}

\noindent
{\bf Feedback on message loss.}  
The \texttt{Human-Message-Loss} $(\widehat{\bm{T}})$ 
solicits user feedback on whether  
template $\widehat{\bm{T}}$ has message loss or not.

\noindent
{\bf Feedback on dummy tokens.}  
The function 
\texttt{Human-Dummy-Token}
$(\widehat{\bm{T}})$ takes template $\widehat{\bm{T}}$, 
which has dummy tokens, 
as input, and it requests the user to select at least one dummy token.  
 
\noindent
{\bf Feedback on message comparison.}  
The function 
\texttt{Human-Select} $(\widehat{\bm{T}}, \mathcal{T})$ 
assists users to select a template from 
the candidate set $\mathcal{T}$ 
that has the same message as the template $\widehat{\bm{T}}$.  
Steps \ref{Step:HumanSelect:LCS} to \ref{Step:HumanSelect:Delete} 
generate a user-friendly list for the user.  
More specifically, this list sorts templates in $\widehat{\bm{T}}$ based 
on their message distance 
(quantified by the length of the longest common subsequence) 
to $\widehat{\bm{T}}$ in descending order.  
Furthermore, this list eliminates templates that share no 
common subsequence with $\widehat{\bm{T}}$.  
The user just needs to scan through the list in order 
to select the one having the same message as  $\widehat{\bm{T}}$.  
The chosen template is returned as the output.  
If none is selected, return ``null''.

\subsection{Message Completion}  

\noindent
{\bf Design objective.} 
Given an extracted cluster-template pair 
$(\widehat{\mathcal{C}}_{k}, \widehat{\bm{T}}_{k})$, 
our objective is to improve the message completeness of 
the template $\widehat{\bm{T}}_{k}$ 
without changing the cluster $\widehat{\mathcal{C}}_{k}$.  
Note that the input $(\widehat{\mathcal{C}}_{k}, \widehat{\bm{T}}_{k})$ 
is specified by the user, which reflects the user's needs or preferences.  
To make the objective more precise, 
we consider the following two cases: 

\begin{itemize}[noitemsep,topsep=0pt,leftmargin=*]
\item {\bf $\widehat{\mathcal{C}}_{k}$ is pure. }  
All logs in cluster $\widehat{\mathcal{C}}_{k}$ 
have the same ground-truth template, 
and we denote this ground-truth template 
as $\bm{T}_{\text{true}} \in \{ \bm{T}_k | k \in [K] \}$.  
Denote the set of all message-complete common subsequence of logs in 
$\widehat{\mathcal{C}}_{k}$ as 
\[
\mathcal{S}_{\text{complete}}  
\triangleq
\left\{
\bm{S} | 
\bm{T}_{\text{true}}
\sqsubseteq 
\bm{S}, 
\bm{S} 
\sqsubseteq 
\bm{L}_n, \forall n \in \widehat{\mathcal{C}}_{k}
\right\}.  
\]
Note that $\bm{T}_{\text{true}} \in \mathcal{S}_{\text{complete}} $, 
i.e., the ground truth template is one element of $\mathcal{S}_{\text{complete}} $.    
Our objective is to locate one element in $\mathcal{S}_{\text{complete}}$.   
Note that the located element may not be the exact ground truth template; 
instead, it may contain some dummy tokens.  
This relaxation of the searching objective 
enables us to design fast algorithms.  
From a practice point of view, 
dummy tokens do not damage the message of a template 
provided that the temple has no message loss.  

\item {\bf $\widehat{\mathcal{C}}_k$ is mixed.} 
Different logs in $\widehat{\mathcal{C}}_k$ may have different ground-truth template.  
Denote the set of all common subsequence of logs in 
$\widehat{\mathcal{C}}_{k}$ that have no less message than $\widehat{\bm{T}}_{k}$ as 
\[
\mathcal{S}_{\text{partial}} 
\triangleq
\left\{
\bm{S} | 
\widehat{\bm{T}}_k 
\sqsubseteq 
\bm{S}, 
\bm{S} 
\sqsubseteq 
\bm{L}_n, \forall n \in \widehat{\mathcal{C}}_{k}
\right\}.  
\]
In general, templates in $\mathcal{S}_{\text{partial}}$ have partial message,  
but they have at least the same message as $\widehat{\bm{T}}_{k}$.  
Our objective is to locate one template in $\mathcal{S}_{\text{partial}}$.   
\end{itemize}
 
\noindent
{\bf Algorithm design \& analysis.}  
Algorithm \ref{Algo:MesgCompletion} outlines procedures  
to achieve the above objectives.   
Algorithm \ref{Algo:MesgCompletion} only needs one pass of the 
logs in $\widehat{\mathcal{C}}_k$ 
and it works in a ``stream'' fashion.  
Steps \ref{Step:MesgCompletion:InitialIndex} and \ref{Step:MesgCompletion:InitialTemplate} 
get one log from cluster $\widehat{\mathcal{C}}_k$.      
It is used to initialize the temporal template, which will be updated later.  
Each iteration in the while loop 
processes one log from the cluster $\widehat{\mathcal{C}}_k$ 
till all logs are processed.  
For each log, 
if the temporal template matches it (step \ref{Step:MesgCompletion:match}), 
i.e, being a subsequence of the log, 
then we move to the next iteration.  
If it does not match the log, 
the longest common subsequence between the temporal template 
and the log is extracted (\ref{Step:MesgCompletion:LCS}).  
The extracted longest common subsequence replaces 
the temporal template (\ref{Step:MesgCompletion:LCS}).   
Early termination happens once 
the temporal template does not have more 
messages than the mined template $\bm{T}_k$ 
(steps \ref{Step:MesgCompletion:LossIf}-\ref{Step:MesgCompletion:LossEnd}).   
The following theorems prove the correctness of Algorithm \ref{Algo:MesgCompletion}.   

\begin{algorithm}[htb]
\caption{\texttt{Message-Completion} $(\widehat{\mathcal{C}}_k, \widehat{\bm{T}}_k)$ } 
\label{Algo:MesgCompletion}
\begin{algorithmic}[1]
\STATE 
$n \gets$ an index from $ \widehat{\mathcal{C}}_k$ 
\label{Step:MesgCompletion:InitialIndex}
\STATE 
$\bm{T}_{\text{mc}} \leftarrow \bm{L}_n$, 
$\widehat{\mathcal{C}}_k \leftarrow \widehat{\mathcal{C}}_k \setminus \{n\}$
\label{Step:MesgCompletion:InitialTemplate}
\STATE
\textbf{while} $ \widehat{\mathcal{C}}_k \neq \emptyset$ \textbf{do}
\STATE
\ \ \ \ \ \
$n \gets$ an index from $ \widehat{\mathcal{C}}_k$
\STATE
\ \ \ \ \ \
\textbf{if} \texttt{Match}($\bm{T}_{\text{mc}}$, $\bm{L}_n$) == 1 \textbf{then}
\label{Step:MesgCompletion:match}
\STATE
\ \ \ \ \ \
\ \ \ \ \ \
$\widehat{\mathcal{C}}_k \leftarrow \widehat{\mathcal{C}}_k \setminus \{n\}$ 
\STATE
\ \ \ \ \ \
\textbf{else}
\STATE
\ \ \ \ \ \
\ \ \ \ \ \
$\bm{T}_{\text{mc}} 
\leftarrow 
\texttt{LCS} 
(\bm{T}_{\text{mc}}, \bm{L}_n)$ 
\label{Step:MesgCompletion:LCS}
\STATE
\ \ \ \ \ \
\ \ \ \ \ \
\textbf{if} \texttt{Match}$(\widehat{\bm{T}}_k, \bm{T}_{\text{mc}}) \neq 1$ \textbf{then}
\label{Step:MesgCompletion:LossIf}
\STATE
\ \ \ \ \ \
\ \ \ \ \ \
\ \ \ \ \ \
$\bm{T}_{\text{mc}} \gets \widehat{\bm{T}}_k$
\STATE
\ \ \ \ \ \
\ \ \ \ \ \
\ \ \ \ \ \
\textbf{Break while}
\label{Step:MesgCompletion:LossEnd}
\STATE
\textbf{return} $\bm{T}_{\text{mc}}$
\STATE
\textbf{SubFunction} \texttt{LCS} ($\bm{a},\bm{b}$)
\STATE
\ \ \ \ \ \
\textbf{return} Longest common subsequence of $\bm{a}$ and $\bm{b}$ \cite{wagner1974string}
\STATE
\textbf{SubFunction} \texttt{Match} ($\bm{a},\bm{b}$) (adapt from \cite{chang1994sublinear})
\end{algorithmic}
\end{algorithm}

\begin{theorem} 
Suppose \texttt{LCS} satisfies that for any $i,j \in \mathcal{C}_k, \forall k$, 
\begin{align}
\bm{T}_k 
\sqsubseteq
\texttt{LCS} ( \bm{L}_i, \bm{L}_j ). 
\label{eq:No-Loss-LCS}
\end{align}
Suppose  $\widehat{\mathcal{C}}_k$ is pure and its associated ground-truth template is $\bm{T}$.  
The output of Algorithm \ref{Algo:MesgCompletion} satisfies that 
$\bm{T}_{\text{mc}} \in \mathcal{S}_{\text{complete}}$ if $\widehat{\mathcal{C}}_k$ is pure,  
otherwise $\bm{T}_{\text{mc}} \in \mathcal{S}_{\text{partial}}$.  
\label{thm:MC:correctness}
\end{theorem}

\noindent
All proofs are in our technical report\footnote{\url{https://arxiv.org/abs/2301.12225}}. 
Theorem \ref{thm:MC:correctness} states that under mild assumptions, 
Algorithm \ref{Algo:MesgCompletion} eliminates loss-pure errors.  
In particular, if the cluster is pure, 
Algorithm \ref{Algo:MesgCompletion} outputs a template that has the complete message.  
Otherwise, Algorithm \ref{Algo:MesgCompletion} outputs a template with 
at least the same message as the mined template.  
The condition \ref{eq:No-Loss-LCS} states that the longest subsequence of 
two logs that have the same template summarizes and extracts the complete message of these two logs.  
In fact, experiments on real-world datasets show that 
condition \ref{eq:No-Loss-LCS} is rarely violated.  
If it is violated, one can apply Algorithm \ref{Algo:Human-CSS} 
(whose details are deferred to the last part of this section)
to extract the message complete subsequence.  
\begin{theorem} \label{thm:MC:ComComplex}
The computational complexity of Algo. \ref{Algo:MesgCompletion} is  
$
O( 
| \widehat{\mathcal{C}}_k | \widehat{M}_{\text{max}} 
\\+ 
\widehat{M}^3_{\text{max}}
)
$
where 
$
\widehat{M}_{\text{max}} 
\triangleq 
\max_{ n \in \widehat{\mathcal{C}}_k } \texttt{len} (\bm{L}_n)
$.   
\end{theorem}
\noindent
Theorem \ref{thm:MC:ComComplex} states that 
the computational complexity is linear in the number of input logs 
with a scaling factor of the maximum length of the input log.    
It is cubic in the maximum length of the input log.

\noindent
{\bf No loss template extraction.} 
Algorithm \ref{Algo:Human-CSS} relies on user feedback to 
extract a template, i.e., a common subsequence, from two sequences of tokens.  
The extracted template does not have message loss.  
It is highly likely that the longest common subsequence of two sequences 
does not have message loss.  
Step \ref{Step:Human-CSS:LCS} extracts the 
longest common subsequence.  
To avoid the rare corner case that the longest common subsequence 
has message loss, 
the user provides feedback on whether the message is complete.   
If not, it indicates that the extracted template must contain some variables.  
In Step \ref{Step:Human-CSS:VariableSelect}, 
the user selects at least one variable out.  
Step \ref{Step:Human-CSS:TrimA} and \ref{Step:Human-CSS:TrimB} 
trim the selected variables from two sequences.  
Steps \ref{Step:Human-CSS:LCSLoop} extracts the longest 
common subsequence between these updated sequences.  
We repeat this process, until the termination condition is met.  

\begin{algorithm}[htb]
\caption{\texttt{Lossless-Template} $(\bm{a}, \bm{b})$} 
\label{Algo:Human-CSS}
\begin{algorithmic}[1]
\STATE
$\widehat{\bm{T}} \gets \texttt{LCS} (\bm{a}, \bm{b})$
\label{Step:Human-CSS:LCS}
\STATE
\textbf{while}
\texttt{Human-Message-Loss}$(\widehat{\bm{T}}) {==}1$ \& $\bm{a} \!\neq$null \& $\bm{b} \!\neq$null
\textbf{do}
\STATE
\ \ \ \ \ \
$\mathcal{V} \gets \texttt{Human-Dummy-Token} ( \widehat{\bm{T}} )$
\label{Step:Human-CSS:VariableSelect}
\STATE
\ \ \ \ \ \
$\bm{a} \gets $ trim elements in $\mathcal{V}$ from $\bm{a}$
\label{Step:Human-CSS:TrimA}
\STATE
\ \ \ \ \ \
$\bm{b} \gets $ trim elements in $\mathcal{V}$ from $\bm{b}$
\label{Step:Human-CSS:TrimB}
\STATE
\ \ \ \ \ \
$\widehat{\bm{T}} \gets \texttt{LCS} (\bm{a}, \bm{b})$  
\label{Step:Human-CSS:LCSLoop}
\STATE  
\textbf{return}
$\widehat{\bm{T}}$
\end{algorithmic}
\end{algorithm}

\noindent
The following lemma derives an upper bound on 
the number of iterations taken by Algorithm \ref{Algo:Human-CSS}.   
It also states the condition under which the output of 
Algorithm \ref{Algo:Human-CSS} has the complete message.  
\begin{lemma}
Algorithm \ref{Algo:Human-CSS} terminates in at most $\min\{ \texttt{len}(\bm{a}), \texttt{len}(\bm{a}) \}$ rounds.  
If $\bm{a}$ and $\bm{b}$ have the same ground-truth template denoted by $\bm{T}$ 
and the user does make errors in providing feedback, 
the output $\widehat{\bm{T}}$ of Algorithm \ref{Algo:Human-CSS} satisfies 
$\bm{T} \sqsubseteq \widehat{\bm{T}}$.
\label{lem:LosslessTemplate}
\end{lemma}


\section{Eliminating Complete-Partial Error}

 \subsection{Design Objective}  
   
Given a set of mined cluster-template pairs 
$
\mathcal{P}_{\text{mg}} 
\subseteq
\{
(\widehat{\mathcal{C}}_k, \widehat{\bm{T}}_k) 
| 
k \in [\widehat{K}]
\}
$ 
our objective is to eliminate the 
complete-partial error in it, 
i.e, merge partial clusters that belong to 
the same ground-truth cluster together.  
Note that the input set $\mathcal{P}_{\text{mg}} $ 
is specified by the user, which reflects the user's needs or preferences.  
To make the objective more precise, 
we consider the following two cases:  

\begin{itemize}[noitemsep,topsep=0pt,leftmargin=*]
\item 
{\bf Clusters with message-loss templates.}  
The associated mixed cluster may cause the message loss of a template, 
or the associated cluster is pure, 
but the base log mining algorithm misses some messages.  
From the user's perspective, it is difficult for them to 
tell whether a cluster is pure or mixed when the associated 
template has message loss.  
Thus, we only aim to identify the message-loss template.  
\item 
{\bf Clusters with message-complete templates.}  
We first define the equivalence between two message-complete templates.   
Two mined templates $\widehat{\bm{T}}_k$ and $\widehat{\bm{T}}_j$ 
are equal with respect to the message 
(denoted by $\widehat{\bm{T}}_k \overset{\text{msg}}{=} \widehat{\bm{T}}_j$), 
if they are message complete and satisfy   
\[
\left( \arg_{ k \in [K] } \bm{T}_k \sqsubseteq \widehat{\bm{T}}_k \right)
=
\left( \arg_{ k \in [K] } \bm{T}_k \sqsubseteq \widehat{\bm{T}}_j \right).  
\] 
Note that the clusters corresponding to two equal templates are partial 
and they belong to the same ground-truth cluster.  
This implies that they should be clustered together.  
We aim to identify such partial clusters out and merge them together.  
\end{itemize}

\subsection{Algorithm Design \& Analysis.} 
Algorithm \ref{Algo:merge} outlines procedures  
to achieve the above merge objectives.   
Algorithm \ref{Algo:merge} only needs one pass of the 
cluster-template pairs in $\mathcal{P}_{\text{mg}} $ and it works in a ``stream'' fashion.  
It maintains a set of the latest distinct cluster-template pairs with the complete message, 
and the set is initialized as an empty set (step \ref{Step:merge:complete}).   
Each iteration of the while loop process on the template-cluster pair 
from $\mathcal{P}_{\text{mg}} $, 
and terminates till all pairs are processed (step \ref{step:merge:while}).  
When a template comes in, the algorithm first searches from the message-complete pairs to 
see whether there exists a message-complete template that 
is a subsequence of the coming template (step \ref{step:merge:match}).  
If a matched one is found, the coming cluster-template pair is 
added to the message-complete set  
(steps \ref{step:merge:matchYes}-\ref{step:merge:matchAdd}).    
If none is found, then request the user to judge whether 
the message is complete.  
If it has message loss, 
add this template and the corresponding cluster to the message loss set (steps \ref{step:merge:loss}-\ref{stem:merge:losscluster}).  
If it has the complete message, 
then we request the user  to select the template that should be merged with this template 
(step \ref{step:merge:select}).    
If none is selected, we add this coming cluster-template pair is 
added to the message-complete set  
(step \ref{step:merge:nonNewClus}).  
If one is selected, 
we add the index of this log to the cluster of the selected template, 
and we replace the selected template by 
the common sequence of the template and the log (steps \ref{step:merge:merge}-\ref{step:merge:mergeupdate}).  

\begin{algorithm}[htb]
\caption{$\texttt{Merge} (\mathcal{P}_{\text{mg}}) $} 
\label{Algo:merge}
\begin{algorithmic}[1]
\STATE
$\mathcal{P}_{\text{loss}} \gets \emptyset$, $\mathcal{P}_{\text{complete}} \gets \emptyset$
\label{Step:merge:complete}
\STATE
\textbf{while} $\mathcal{P}_{\text{mg}} \neq \emptyset$ \textbf{do}
\label{step:merge:while}
\STATE  
\ \ \ \ \ \
$(\widehat{\bm{T}}, \widehat{\mathcal{C}}) \gets $ a template-cluster pair form $\mathcal{P}_{\text{mg}}$
\STATE  
\ \ \ \ \ \ 
$\mathcal{P}_{\text{mg}} 
\gets 
\mathcal{P}_{\text{mg}} \setminus \{  (\widehat{\bm{T}}, \widehat{\mathcal{C}}) \}$  
\STATE  
\ \ \ \ \ \ 
$(\bm{T}_{\text{match}}, \mathcal{C}_{\text{match}} ) 
\in  
\arg_{ ( \bm{T}, \mathcal{C} ) \in \mathcal{P}_{\text{complete}}} \bm{T} \sqsubseteq \widehat{\bm{T}}$
\label{step:merge:match}
\STATE  
\ \ \ \ \ \
\textbf{if} $(\bm{T}_{\text{match}}, \mathcal{C}_{\text{match}} )  \neq \text{null}$ \textbf{then}
\label{step:merge:matchYes}
\STATE  
\ \ \ \ \ \
\ \ \ \ \ \
$\mathcal{P}_{\text{complete}} 
\gets
 \mathcal{P}_{\text{complete}} \setminus
\{ (\bm{T}_{\text{match}}, \mathcal{C}_{\text{match}} ) \}
$ 
\STATE  
\ \ \ \ \ \
\ \ \ \ \ \
$\mathcal{P}_{\text{complete}} 
\gets 
\mathcal{P}_{\text{complete}}
\cup
\{ (\bm{T}_{\text{match}}, \mathcal{C}_{\text{match}}  \cup \widehat{\mathcal{C}}) \}$
\label{step:merge:matchAdd}
\STATE  
\ \ \ \ \ \
\textbf{else}
\STATE  
\ \ \ \ \ \
\ \ \ \ \ \
\textbf{if} \texttt{Human-Message-Loss} $( \widehat{\bm{T}}) ==1$ \textbf{then}
\label{step:merge:loss}
\STATE  
\ \ \ \ \ \
\ \ \ \ \ \
\ \ \ \ \ \
$\mathcal{P}_{\text{loss}} 
\gets
\mathcal{P}_{\text{loss}} 
\cup 
\{  (\widehat{\bm{T}}, \widehat{\mathcal{C}}) \}
$ 
\label{stem:merge:losscluster}
\STATE  
\ \ \ \ \ \
\ \ \ \ \ \
\textbf{else}
\STATE  
\ \ \ \ \ \
\ \ \ \ \ \
\ \ \ \ \ \
$\mathcal{T}_{\text{complete}} 
\gets
\{
\bm{T} 
|
(\bm{T}, \mathcal{C}) \in \mathcal{P}_{\text{complete}}
\}
$
\STATE  
\ \ \ \ \ \
\ \ \ \ \ \
\ \ \ \ \ \
$\bm{T}_{hs}=$ \texttt{Human-Select}$(\widehat{\bm{T}},\mathcal{T}_{\text{complete}})$
\label{step:merge:select}
\STATE  
\ \ \ \ \ \
\ \ \ \ \ \
\ \ \ \ \ \
\textbf{if} $\bm{T}_{hs}$ == null \textbf{then}
\label{step:merge:non}
\STATE  
\ \ \ \ \ \
\ \ \ \ \ \
\ \ \ \ \ \
\ \ \ \ \ \
$\mathcal{P}_{\text{complete}} 
\gets 
\mathcal{P}_{\text{complete}} \cup \{ (\widehat{\bm{T}}, \widehat{\mathcal{C}}) \}$ 
\label{step:merge:nonNewClus}
\STATE  
\ \ \ \ \ \
\ \ \ \ \ \
\ \ \ \ \ \
\textbf{else}
\STATE  
\ \ \ \ \ \
\ \ \ \ \ \
\ \ \ \ \ \
\ \ \ \ \ \
$\bm{T}_{\text{merge}} \gets$\texttt{Lossless-Template} $(\bm{T}_{hs}, \widehat{\bm{T}})$ 
\label{step:merge:merge}
\STATE  
\ \ \ \ \ \
\ \ \ \ \ \
\ \ \ \ \ \
\ \ \ \ \ \
$\mathcal{C}_{hs} \gets 
$
the cluster associated with $\bm{T}_{hs}$
\STATE  
\ \ \ \ \ \
\ \ \ \ \ \
\ \ \ \ \ \
\ \ \ \ \ \
$\mathcal{P}_{\text{complete}} 
\gets
 \mathcal{P}_{\text{complete}} \setminus
\{ (\bm{T}_{\text{hs}}, \mathcal{C}_{\text{hs}} ) \} 
$
\STATE  
\ \ \ \ \ \
\ \ \ \ \ \
\ \ \ \ \ \
\ \ \ \ \ \
$
\mathcal{P}_{\text{complete}} 
\gets
 \mathcal{P}_{\text{complete}}
\cup 
\{ (\bm{T}_{\text{hs}}, \mathcal{C}_{\text{hs}} \cup \widehat{\mathcal{C}}) \}
$ 
\label{step:merge:mergeupdate}
\STATE
\textbf{return} $\mathcal{P}_{\text{loss}}$ and $\mathcal{P}_{\text{complete}}$
\end{algorithmic}
\end{algorithm}

\begin{theorem} 
\label{thm:merge:correctness}  
Suppose the user provides correct feedback.  
The output of Algorithm \ref{Algo:merge}, i.e., $\mathcal{P}_{\text{loss}}$ and $\mathcal{P}_{\text{complete}}$, satisfies: 
\begingroup
\allowdisplaybreaks
\begin{align}
& 
\cup_{ ( \mathcal{C}, \bm{T} ) \in  \mathcal{P}_{\text{complete}} \cup \mathcal{P}_{\text{loss}} } 
\mathcal{C}
=
\cup_{ ( \mathcal{C}, \bm{T} ) \in \mathcal{P}_{\text{mg}} } \mathcal{C}   
\label{eq:Merge:AllHandle}
\\
& 
\mathcal{P}_{\text{loss}} 
= 
\left\{
(\mathcal{C}, \bm{T}) 
| 
(\mathcal{C}, \bm{T}) \in \mathcal{P}_{\text{mg}}, 
\bm{T} \text{has message loss}
\right\},
\label{eq:Merge:loss}
\\
&
\{ 
(\mathcal{C}, \bm{T})
| 
(\mathcal{C}, \bm{T}) \in \mathcal{P}_{\text{complete}}, 
\bm{T} \text{has message loss}
\}=\emptyset. 
\label{eq:Merge:Completeloss}
\\
&
\{ 
(\mathcal{C}, \bm{T})
| 
(\mathcal{C}, \bm{T}) \in \mathcal{P}_{\text{complete}}, 
\mathcal{C} \text{is mixed}
\}=\emptyset. 
\label{eq:Merge:CompleteMix}
\\
& 
\nexists 
( \widehat{\mathcal{C}}_k, \widehat{\bm{T}}_k ), 
( \widehat{\mathcal{C}}_j, \widehat{\bm{T}}_j ) 
\in \mathcal{P}_{\text{complete}}, 
\widehat{\bm{T}}_k \overset{\text{msg}}{=} \widehat{\bm{T}}_j
\label{eq:Merge:CompleteRedundancy}
\end{align} 
\endgroup
 \end{theorem}
\noindent
Theorem \ref{thm:merge:correctness} states 
that each log is placed either in the message-loss set or the message-complete set 
(Eq. (\ref{eq:Merge:AllHandle})).  
The message-loss set contains exactly all cluster-templates pair with message loss   
(Eq. (\ref{eq:Merge:loss})).  
All pure clusters with complete messages are properly merged 
whenever needed and placed in the message-complete set 
such that all clusters in the message-complete set are pure 
with a template having the complete message, 
and there does not exist two templates 
equal respect to the message (Eq. (\ref{eq:Merge:Completeloss})-(\ref{eq:Merge:CompleteRedundancy})).  

We define the following notations to quantify the number 
of complete-message templates and 
the number of distinct (in the sense of message equivalence) complete-message templates 
in the input set $\mathcal{P}_{\text{mg}}$: 
$
N^{\text{dst}}_{\text{mg}} 
\triangleq 
|
\{
k 
| 
k \in [K], 
\exists (\mathcal{C}, \bm{T}) 
\in 
\mathcal{P}_{\text{mg}}, 
\bm{T}_k \sqsubseteq \bm{T}  
\}
|.  
$
 
\begin{theorem} \label{thm:Merge:computational}
Suppose the user provides correct feedback.  
The computational complexity of Algo. \ref{Algo:merge} can be derived as  
$
O
\Big( 
|\mathcal{P}_{\text{mg}}| 
N^{\text{dst}}_{\text{mg}}
\widetilde{M}_{\text{max}} 
+ 
 ( \widetilde{d}_{\text{max}} +1) 
\left(
N^{\text{dst}}_{\text{mg}}
\right)^2 
\big(\ln N^{\text{dst}}_{\text{mg}}
+
\widetilde{M}^2_{\text{max}}
\big) 
+ 
N^{\text{dst}}_{\text{mg}}  
\widetilde{d}^2_{\text{max}}
\widetilde{M}^2_{\text{max}}
\Big)
$, where 
$
\widetilde{M}_{\text{max}} 
\triangleq 
\max_{ (\mathcal{C}, \bm{T}) \in  \mathcal{P}_{\text{mg}} }  
\text{\# of tokens in } \bm{T}.  
$
\end{theorem}
 
\noindent 
Theorem \ref{thm:Merge:computational} states that the computational complexity of 
Algorithm \ref{Algo:merge} 
is linear in the number number of input templates, 
with a scaling factor of the number of distinct templates in the input 
and the maximum length of the templates in the input.  
It is quadratic in both the maximum length of the templates in the input and 
the maximum number of dummy tokens of templates in the input 
and roughly quadratic in the number of distinct templates 
in the input.  

\begin{theorem} \label{thm:Merge:Query}
Suppose the user provides correct feedback.  
The number of user feedback requested by Algo. \ref{Algo:merge} satisfies:  
\begin{align*}
& 
\# \text{ of } \texttt{Human-Message-Less} \text{ feedback }
= 
|\mathcal{P}_{mg}|,
\\
& 
\# \text{ of } \texttt{Human-Select} \text{ feedback } 
= 
N^{\text{dst}}_{\text{mg}}  ( \widetilde{d}_{\text{max}} +1),
\\
& 
\# \text{ of } \texttt{Human-Dummy-Token} \text{ feedback }  
\leq 
N^{\text{dst}}_{\text{mg}}  
\widetilde{d}^2_{\text{max}},
\end{align*}
where $
\widetilde{d}_{\text{max}}
\triangleq
\max_{ (\mathcal{C}, \bm{T}) \in  \mathcal{P}_{\text{mg}} }  
\bm{1}_{ \{ \bm{T} \text{ has complete message} \} } 
\big( 
\texttt{len} ( \bm{T} ) 
- 
$
\\
$
\texttt{len} 
(
\text{ground-truth template of } \mathcal{C} 
)
\big).  
$
\end{theorem} 
 
\noindent
Theorem \ref{thm:Merge:Query} states that 
the number of human-message-loss feedback requested by Algo. \ref{Algo:merge} 
is exactly the number of cluster-template pairs in the input,  
while the numbers of human-select or human-dummy-token feedback requested by Algo. \ref{Algo:merge} 
are invariant of the number of cluster-template pairs in the input.   
In particular, the number of human-select feedback 
and increases linearly in the number of distinct message-complete 
templates in the input and 
increases linearly in the maximum number of dummy tokens of input templates.    
The number of human-dummy-token feedback 
increases linearly in the number of distinct message-complete 
templates in the input 
increases quadratically in the maximum number of dummy tokens of input templates.


 
 \section{Eliminating Loss-Mixed Error}

\subsection{Design Objective}

We design an algorithm to eliminate loss-mixed error.  
Users rely on the message of a template to assess whether a 
cluster is mixed or pure.  
A pure cluster may be input for separation.    
In this case, our objective is: 
Separate different clusters out, 
and for each separate clusters extract its template with the complete message.  
We want to emphasize that the message is complete, 
not meaning the exact template.  
This also works for the case that the input cluster 
is pure.  

\subsection{ Algorithm Design \& Analysis}  
Algorithm \ref{Algo:Separation} outlines our algorithm to eliminate loss-mixed error.  
It only needs to conduct one pass of the logs in cluster $\widehat{\mathcal{C}}_{\text{sep}}$ 
in a stream fashion.  
It maintains a set of the latest distinct cluster-template, 
and the list is initialized as an empty list (step \ref{Step:Separation:Initial}).   
When a log comes in, the algorithm first search from the template list to 
see whether there exists a template that matches the log (steps \ref{Step:Separation:Index}-\ref{Step:Separation:Matching}).  
If a matched one is found, add the index of the log to the matched template cluster (step \ref{Step:Separation:Matched} and \ref{Step:Separation:MatchedAdd}).  
If none is found, sort templates in the list 
and request the user to select a template from the candidate set 
that should be merged with this log (step \ref{Step:Separation:Select}).  
If none is selected, we add this log to the template list 
and initialize its associated cluster as the index of the log (\ref{Step:Separation:InitialTemplate}).  
If one is selected, 
we add the index of this log to the cluster of the selected template, 
and we replace the selected template by 
the common sequence of the template and the log (steps \ref{Step:Separation:GetTemplate}-\ref{Step:Separation:Update}).

\begin{algorithm}[htb]
\caption{Human-in-the-loop Template Mining} 
\label{Algo:Combination} 
\begin{flushleft} 
{\bf Input:} a set of logs $\{ \bm{L}_n | n \in [N] \}$, $N_{\text{repeat}}$
\\ \hspace{0.35in} base template mining algorithm $\texttt{BaseAlgo}$  \\
{\bf Output:} a set of cluster-template pairs $\mathcal{P}$
\end{flushleft}
\begin{algorithmic}[1]
\STATE
$
\mathcal{P}_{\text{temp} }
\gets 
\texttt{BaseAlgo} (\{ \bm{L}_n | n \in [N] \})
$,\quad
$\mathcal{Q}_{\text{temp}} \gets \emptyset$

\STATE 
\textbf{while} $\mathcal{P}_{\text{temp}} \neq \emptyset$ \textbf{do}
\STATE
\ \ \ \ \ \
$(\mathcal{C}, \bm{T}) \gets$ a cluster-template pair from $\mathcal{P}_{\text{temp}}$  
\STATE 
\ \ \ \ \ \
$
\mathcal{P}_{\text{temp}} 
\gets 
\mathcal{P}_{\text{temp}} 
\setminus 
\{ (\mathcal{C}, \bm{T}) \}
$
\STATE
\ \ \ \ \ \
$\bm{T} \gets \texttt{Message-Completion} (\mathcal{C}, \bm{T})$  
\STATE 
\ \ \ \ \ \
$\mathcal{Q}_{\text{temp}} \gets 
\mathcal{Q}_{\text{temp}} 
\cup 
\{ (\mathcal{C}, \bm{T}) \}
$

\STATE
\textbf{while} $N_{\text{repeat}} \geq 0$ \textbf{do}
\STATE
\ \ \ \ \ \
$
N_{\text{repeat}}
\gets 
N_{\text{repeat}} - 1
$,
\quad
$( \mathcal{P}_{\text{loss}},\mathcal{P}_{\text{complete}}) 
\gets 
\texttt{Merge} (\mathcal{Q}_{\text{temp}})
$

\STATE 
\ \ \ \ \ \
\textbf{if} $ \mathcal{P}_{\text{loss}} ==\emptyset$ \textbf{then}
\STATE 
\ \ \ \ \ \
\ \ \ \ \ \
{\bf Break while}

\STATE 
\ \ \ \ \ \
\textbf{while} $\mathcal{P}_{\text{loss}} \neq \emptyset$ \textbf{do}

\STATE 
\ \ \ \ \ \
\ \ \ \ \ \
$(\widehat{C}_{\text{sep}}, \widehat{\bm{T}}_{\text{sep}}) 
\gets$ one cluster-template pair from $\mathcal{P}_{\text{loss}} $

\STATE 
\ \ \ \ \ \
\ \ \ \ \ \
$\mathcal{P}_{\text{sep}} 
\gets 
\texttt{Separation} (\widehat{C}_{\text{sep}})
$

\STATE 
\ \ \ \ \ \
\ \ \ \ \ \
$\mathcal{P}_{\text{complete}}
\gets 
\mathcal{P}_{\text{complete}} 
\cup 
\mathcal{P}_{\text{sep}}
$
\STATE 
\ \ \ \ \ \
$
\mathcal{Q}_{temp}  \gets \mathcal{P}_{\text{complete}}
$
\STATE
$\mathcal{P} \gets \mathcal{Q}_{temp} $
\RETURN $\mathcal{P}$
\end{algorithmic}
\end{algorithm}

\begin{theorem}
\label{thm:separation:correct}
Suppose the user provide correct feedback.  
The output of Algorithm \ref{Algo:Separation}, i.e., $\mathcal{P}_{\text{sep}}$, 
satisfies: 
\begingroup
\allowdisplaybreaks   
\begin{align}
& 
\cup_{ (\mathcal{C}, \bm{T}) \in \mathcal{P}_{\text{sep}} }  \mathcal{C} 
= 
\widehat{\mathcal{C}}_{\text{sep}},
\label{eq:Separation:AllHandle}
\\
&
\{ 
(\mathcal{C}, \bm{T})
| 
(\mathcal{C}, \bm{T}) \in \mathcal{P}_{\text{sep}}, 
\bm{T} \text{has message loss}
\}=\emptyset, 
\label{eq:Separation:Completeloss}
\\
&
\{ 
(\mathcal{C}, \bm{T})
| 
(\mathcal{C}, \bm{T}) \in \mathcal{P}_{\text{sep}}, 
\mathcal{C} \text{is mixed}
\}=\emptyset,
\label{eq:Separation:CompleteMix}
\\
& 
\nexists 
( \widehat{\mathcal{C}}_k, \widehat{\bm{T}}_k ), 
( \widehat{\mathcal{C}}_j, \widehat{\bm{T}}_j ) 
\in \mathcal{P}_{\text{sep}}, 
\widehat{\bm{T}}_k \overset{\text{msg}}{=} \widehat{\bm{T}}_j.
\label{eq:Separation:CompleteRedundancy}
\end{align} 
\endgroup
\end{theorem}

\noindent
Theorem \ref{thm:merge:correctness} states that 
the output of Algorithm \ref{Algo:Separation} is correct when 
the user provides correct feedback.   
Specifically, each log in the input belongs to 
one of the templates in the output  
(Eq. (\ref{eq:Separation:AllHandle})).  
All clusters in the output set are pure 
with templates having the complete message, 
and there do not exist two templates 
with equal respect to the message (Eq. (\ref{eq:Separation:Completeloss})-(\ref{eq:Separation:CompleteRedundancy})).  

\begin{theorem} \label{thm:Separation:QueryComp}
Suppose the user provides correct feedback.  
The number of user feedback requested by Algo. \ref{Algo:Separation} satisfies: 
\begin{align*} 
& 
\# \text{ of } \texttt{Human-Select} \text{ feedback }
\leq 
N^{\text{tpl}}_{\text{sep}}   
(d_{\text{max}} +1), 
\\
& 
\# \text{ of } \texttt{Human-Dummy-Token} \text{ feedback }
\leq 
N^{\text{tpl}}_{\text{sep}} d^2_{\text{max}},   
\end{align*}
where 
$
N^{\text{tpl}}_{\text{sep}} 
\triangleq 
| 
\{
k 
|
k \in [K], 
\exists 
n \in \widehat{\mathcal{C}}_{\text{sep}}, 
\bm{T}_k \sqsubseteq \bm{L}_n
\}
|
$ 
and 
$d_{\text{max}} 
\triangleq 
\max_{ n \in \widehat{\mathcal{C}}_{\text{sep}}} 
M_n 
- 
\texttt{len} ( \text{template of } \bm{L}_n)
$.  
\end{theorem}

\noindent
Theorem \ref{thm:Separation:QueryComp} states that 
the number of user feedback requested by Algo. \ref{Algo:Separation} 
is invariant with the number of input logs.   
In particular, the number of human-select feedback 
increases linearly in the number of distinct templates associated 
with the input logs and 
increases linearly in the maximum number of dummy tokens of logs.  
In particular, the number of human-dummy-token feedback 
increases linearly in the number of distinct templates associated 
with the input logs and 
increases quadratically in the maximum number of dummy tokens of logs.

\begin{theorem} \label{thm:Separation:Comp}
The computational complexity of Algorithm \ref{Algo:Separation} can be derived as 
$
O \Big(
| 
\widehat{\mathcal{C}}_{\text{sep}}
|  
N^{\text{tpl}}_{\text{sep}}
M_{\text{max}}
+ 
(d_{\text{max}} +1) 
\left( N^{\text{tpl}}_{\text{sep}} \right)^2
\left( 
\ln 
N^{\text{tpl}}_{\text{sep}}
+ 
M^2_{\text{max}}
\right)
+
N^{\text{tpl}}_{\text{sep}} d^2_{\text{max}} 
M^2_{\text{max}}
\Big)
$, 
where $M_{\text{max}} \triangleq \max_{n \in \widehat{\mathcal{C}}_{\text{sep}} } M_n$.  
\end{theorem}

\noindent 
Theorem \ref{thm:Separation:Comp} states that the computational complexity 
is linear in the size of input $\widehat{\mathcal{C}}_{\text{sep}}$, 
with a scaling factor of the number of distinct templates in the input 
and the maximum length of the logs in the input.  
It is quadratic in both the maximum length of the logs in the input and 
the maximum number of dummy tokens of logs in the input 
and roughly quadratic in the number of distinct templates 
associated with the input.

\section{Applications}

Engineers can apply our proposed 
\texttt{Message-Completion}, \texttt{Merge}, and \texttt{Separation} 
separately or combine some of them to fulfill their needs.  
Here, Algorithm \ref{Algo:Combination}  shows 
one combination of these algorithms, 
which is generic to improve the accuracy of any based  
template mining algorithms.  
Algorithm \ref{Algo:Combination} first applies the base template 
mining algorithm to extract templates of logs.  
Then it applies \texttt{Message-Completion} 
to eliminate pure-loss errors.  
Then it repeats the merge-separation process for a given 
number of $N_{\text{repeat}}$ rounds.  
In each round, it first applies the \texttt{Merge} 
algorithm to eliminate complete-partial errors, 
then applies the \texttt{Separation} algorithm to 
eliminate loss-mixed errors.  
Early termination happens when there are errors.  

\begin{algorithm}[htb]
\caption{ $\texttt{Separation} (\widehat{\mathcal{C}}_{\text{sep}})$ } 
\label{Algo:Separation}
\begin{algorithmic}[1] 
\STATE 
$\mathcal{P}_{sep} \gets \emptyset$
\label{Step:Separation:Initial}

\STATE
\textbf{while} {$\widehat{\mathcal{C}}_{\text{sep}} \neq \emptyset$} \textbf{do}

\STATE 
\ \ \ \ \ \
$n \gets$ an index from $ \widehat{\mathcal{C}}_{\text{sep}}$ 
\label{Step:Separation:Index}

\STATE 
\ \ \ \ \ \
$\widehat{\mathcal{C}}_{\text{sep}} \gets \widehat{\mathcal{C}}_{\text{sep}} \setminus \{ n \}$  

\STATE 
\ \ \ \ \ \
$
(\mathcal{C}_{\text{match}}, \bm{T}_{\text{match}} ) 
\in \arg_{ (\mathcal{C}, \bm{T} ) \in \mathcal{P}_{\text{sep} } }  
\bm{T} \sqsubseteq \bm{L}_n 
$
\label{Step:Separation:Matching}

\STATE 
\ \ \ \ \ \
\textbf{if} $(\mathcal{C}_{\text{match}}, \bm{T}_{\text{match}} ) \neq$ null \textbf{then}
\label{Step:Separation:Matched}

\STATE 
\ \ \ \ \ \
\ \ \ \ \ \
$\mathcal{C}_{\text{match}} \gets 
\mathcal{C}_{\text{match}} \cup \{ n \}
$ 
\label{Step:Separation:MatchedAdd}

\STATE 
\ \ \ \ \ \
\textbf{else}
\STATE 
\ \ \ \ \ \
\ \ \ \ \ \
$\mathcal{T}_{\text{sep}} 
\gets
\{
\bm{T} 
|
(\bm{T}, \mathcal{C}) \in \mathcal{P}_{\text{sep}}
\}
$
\STATE 
\ \ \ \ \ \
\ \ \ \ \ \
$\bm{T}_{\text{hs}} \leftarrow \texttt{Human-Select}(\bm{L}_n,\mathcal{T}_{\text{sep}})$
\label{Step:Separation:Select}

\STATE 
\ \ \ \ \ \
\ \ \ \ \ \
\textbf{if} $\bm{T}_{\text{hs}} == $ null \textbf{then}

\STATE 
\ \ \ \ \ \
\ \ \ \ \ \
\ \ \ \ \ \
$
\mathcal{P}_{\text{sep}} 
\gets 
\mathcal{P}_{\text{sep}}  
\cup
\{ (\{n\}, \bm{L}_n) \}
$ 
\label{Step:Separation:InitialTemplate}

\STATE 
\ \ \ \ \ \
\ \ \ \ \ \
\textbf{else}
\STATE 
\ \ \ \ \ \
\ \ \ \ \ \
\ \ \ \ \ \
$\mathcal{C}_{\text{hs}} 
\gets
$
the cluster associated with 
$\bm{T}_{\text{hs}}$ 
\label{Step:Separation:GetTemplate}

\STATE 
\ \ \ \ \ \
\ \ \ \ \ \
\ \ \ \ \ \
$
\mathcal{P}_{\text{sep}} 
\gets 
\mathcal{P}_{\text{sep}}  
\setminus 
\{ (\mathcal{C}_{\text{hs}}, \bm{T}_{\text{hs}}) \}
$

\STATE 
\ \ \ \ \ \
\ \ \ \ \ \
\ \ \ \ \ \
$\bm{T}_{\text{merge}} \leftarrow \texttt{Lossless-Template} (\bm{T}_{\text{hs}}, \bm{L}_n)$
\label{Step:Separation:LosslessTemplate}

\STATE 
\ \ \ \ \ \
\ \ \ \ \ \
\ \ \ \ \ \ 
$
\mathcal{P}_{\text{sep}} 
\gets 
\mathcal{P}_{\text{sep}}  
\cup
\{ (\mathcal{C}_{\text{hs}} \cup \{n\}, \bm{T}_{\text{merge}}) \}
$
\label{Step:Separation:Update}

\RETURN $\mathcal{P}_{sep}$  
\end{algorithmic}
\end{algorithm}

\section{Experiments}

\subsection{Experiment Setting} 

We conduct experiments on sixteen widely used benchmark datasets \cite{zhu2019tools}.  
We recruit ten graduate students to conduct human-in-the-loop experiments.  
To extensively evaluate our proposed algorithms under a large 
number of settings and datasets, 
Algorithm \ref{Algo:FeedabckSimulator}  
outlines procedures to simulate human feedback.  
If template $\widehat{\bm{T}}$ has message loss, 
the   
\texttt{Simulator-Select} and 
\texttt{Simulator-Dummy-Token} 
always return null.  
Namely, these two simulators are weaker than human.  
The \texttt{Simulator-Message-Loss}  always provides correct feedback.  
This simulator is not weaker than human.  
We consider five popular base template mining algorithms: 
Drain \cite{he2017drain}, Spell \cite{du2016spell}, IPLoM \cite{makanju2009clustering}, 
Logram \cite{dai2020logram}, Prefix Graph \cite{chu2021prefix}.   
For each base algorithm, we consider two 
parameters: 
(1) fine tuned parameter that achieves nearly the best performance 
on each dataset \cite{zhu2019tools, chu2021prefix}; 
(2) an arbitrarily selected sub-optimal parameter.  
In the following, we first evaluate the overall performance of 
a combination of our proposed algorithms, 
i.e., Algorithm \ref{Algo:Combination}.  
Unless explicitly stated otherwise, we set the parameter $N_{\text{repeat}}$ of 
Algorithm \ref{Algo:Combination} as 0, 
i.e., do not repeat.  
Then we evaluate each individual human-in-the-loop algorithm.

\begin{algorithm}[htb]
\caption{\texttt{Feedback Simulator} } 
\label{Algo:FeedabckSimulator} 

\begin{algorithmic}[1]
\STATE
\textbf{SubFunction} 
\texttt{Simulator-Message-Loss} 
({
$\widehat{\bm{T}}$
})
\STATE  
\ \ \ \ \ \
\textbf{return}
$
\bm{I}_{ \{ \{ k | k \in [K], \bm{T}_k \sqsubseteq \widehat{\bm{T}} \} \neq \emptyset \} }
$
\STATE
\textbf{SubFunction} 
\texttt{Simulator-Dummy-Token} ($\widehat{\bm{T}}$)
\STATE  
\ \ \ \ \ \
$k' \gets \{ k | k \in [K], \bm{T}_k \sqsubseteq \widehat{\bm{T}} \}$
\STATE  
\ \ \ \ \ \
$\mathcal{A} \gets \{ a | a \text{ is an element of } \widehat{\bm{T}},  
a \text{ is not an element of } \bm{T}_{k'}  \}$
\STATE  
\ \ \ \ \ \
\textbf{return}
the first element of $\mathcal{A}$
\STATE
\textbf{SubFunction}
\texttt{Simulator-Select} ($\widehat{\bm{T}}, \mathcal{T}$)
\STATE
\ \ \ \ \ \
$\bm{T} \in \arg_{ k \in [K]} \bm{T}_k \sqsubseteq \widehat{\bm{T}}$,
\ \
$\bm{T}' \in 
\arg_{ \widetilde{\bm{T}} \in \mathcal{T} }  
\bm{T} \sqsubseteq  \widetilde{\bm{T}}
$ 
\STATE  
\ \ \ \ \ \
\textbf{return}
$\bm{T}' $
\end{algorithmic}
\end{algorithm}

\subsection{Evaluating the Overall Performance}
\label{sec:exp:overall}

Table \ref{Tb:humanVsSimulator} shows the GA 
and MA of Algorithm \ref{Algo:Combination} 
under human feedback and 
simulated feedback respectively.  
Due to constraints in human resources, 
we only select four datasets to conduct human feedback experiments.  
The column labeled ``human'' (or ``simu.'') denotes the 
accuracy of Algorithm \ref{Algo:Combination}  under human 
(or simulated) feedback.  
One can observe that the GA (or MA) of 
Algorithm \ref{Algo:Combination} is no less than 0.95 (0.99)
under both human feedback and simulated feedback.  
In other words, Algorithm \ref{Algo:Combination} 
has extremely high accuracies.  
Furthermore, the GA (or the MA) 
of Algorithm \ref{Algo:Combination} under the simulated feedback is nearly the same as that under 
human feedback.  
This shows that our simulator is accurate in approximating 
human feedback.  
Thus, in later experiments, we use our simulator to test 
our proposed algorithms under a large number of settings.

\begin{table}[htb]
\caption{Accuracy (human feedback vs. simulator).} 
\label{Tb:humanVsSimulator}
\centering
	\begin{tabular}{p{0.88cm}ccccc}
		\hline       
        \multirow{2}{*}{\textbf{Method}}      & \multirow{2}{*}{\textbf{Dataset}} & \multicolumn{2}{c}{\textbf{GA}}        & \multicolumn{2}{c}{\textbf{MA}}     \\ \cline{3-6} 
&                                   & \textbf{human} & \textbf{simu.} & \textbf{human} & \textbf{simu.} \\                                             
		\hline
		\multirow{4}{*}{Drain}                                                  
		& Android          & 0.998             & 0.998                 & 0.9995            & 0.9995                \\
		& BGL              & 1                 & 1                     & 1                 & 1                     \\
		& HPC              & 1                 & 1                     & 1                 & 1                     \\
		& OpenStack        & 1                 & 1                 & 1                 & 1                     \\
		\hline
		\multirow{4}{*}{Spell}                                                  
		& Android          & 0.998             & 0.998                 & 0.9995            & 0.9995                \\
		& BGL              & 1                 & 1                     & 1                 & 1                     \\
		& HPC              & 0.9579            & 0.9595                & 1                 & 1                     \\
		& OpenStack        & 1                 & 1                 & 1                 & 1                     \\
		\hline
		\multirow{4}{*}{IPLoM}                                                  
		& Android          & 0.998             & 1                 & 0.9995            & 0.9995                \\
		& BGL              & 1                 & 1                     & 1                 & 1                \\
		& HPC              & 1                 & 1                 & 1                 & 1                     \\
		& OpenStack        & 1                 & 1                     & 1                 & 1                     \\
		\hline
		\multirow{4}{*}{Logram}                                             
		& Android          & 0.998             & 0.998                 & 0.9995            & 0.9995                \\
		& BGL              & 1                 & 1                     & 1                 & 1                     \\
		& HPC              & 0.962             & 0.9665                & 1                 & 1                     \\
		& OpenStack        & 1                 & 1                     & 1                 & 1                     \\
		\hline
		\multirow{4}{*}{\begin{tabular}[c]{@{}c@{}}Prefix\\ Graph\end{tabular}} 
		& Android          & 0.998             & 0.998                 & 0.9995            & 0.9995                \\
		& BGL              & 1                 & 1                 & 1                 & 1                \\
		& HPC              & 1                 & 1                     & 1                 & 1                     \\
		& OpenStack        & 1                 & 1                     & 1                 & 1 \\ 
		\hline
	\end{tabular}
\end{table}

Table \ref{Tb:Accuracy:DrainFine} shows the accuracy improvement 
of Algorithm \ref{Algo:Combination} over the base template mining algorithm Drain.  
We run Algorithm \ref{Algo:Combination} with simulated feedback 
and run Drain with fine tuned parameters.  
The column labeled ``drain'' (or ``simu.'' or ``rpt'') denotes the 
accuracy of the base template mining algorithm Drain 
(or Algorithm \ref{Algo:Combination} without repeat or 
Algorithm \ref{Algo:Combination}  with one round of $N_{\text{repeat}}=1$ ). 
One can observe that Algorithm \ref{Algo:Combination} improves the GA of Drian 
to close to 1, 
whether the GA under Drain is high or low.  
Repeat our Algorithm \ref{Algo:Combination} to do another round of merge and separation; 
the GA is increased to nearly 1.  
Similar observations can be found on the MA metric.  
Table \ref{Tb:Accuracy:DrainSubOpt}  shows similar improvement in GA and MA 
when Drain is run with sub-optimal parameters.  
They show the superior performance of Algorithm \ref{Algo:Combination}.  
For other base template mining algorithms, i.e., Spell, IPLoM, etc., 
Algorithm \ref{Algo:Combination} has a similar improvement in accuracy.  
Due to the page limit, more experiments are in the appendix.  
\begin{table}[htb]
	\caption{Accuracy (Drain, fine tuned parameter).}  
	\label{Tb:Accuracy:DrainFine}
	\centering
	\begin{tabular}{p{1.1cm}cccccc}
		\hline
		\multirow{2}{*}{\textbf{Dataset}} & \multicolumn{3}{c}{\textbf{GA}}                    & \multicolumn{3}{c}{\textbf{MA}}                    \\ \cline{2-7} 
& \textbf{drain} & \textbf{simu.} & \textbf{rpt} & \textbf{drain} & \textbf{simu.} & \textbf{rpt} \\
		\hline
		Andriod     & 0.911     & 0.998    & 0.998     & 0.972     & 0.9995   & 0.9995    \\
		Apache      & 1         & 1        & 1         & 1         & 1        & 1         \\
		BGL         & 0.9625    & 1        & 1         & 0.976     & 1        & 1         \\
		Hadoop      & 0.9475    & 0.9975   & 1         & 0.963     & 1        & 1         \\
		HDFS        & 0.9975    & 1        & 1         & 1         & 1        & 1         \\
		HealthApp   & 0.78      & 1        & 1         & 0.9005    & 1        & 1         \\
		HPC         & 0.887     & 1        & 1         & 0.8965    & 1        & 1         \\
		Linux       & 0.69      & 0.8785   & 1         & 0.7515    & 0.941    & 0.941     \\
		Mac         & 0.7865    & 0.902    & 1         & 0.907     & 0.99     & 0.99      \\
		OpenSSH     & 0.7875    & 1        & 1         & 0.7865    & 1        & 1         \\
		OpenStack   & 0.7325    & 0.989    & 1         & 0.207     & 1        & 1         \\
		Proxifier   & 0.5265    & 1        & 1         & 1         & 1        & 1         \\
		Spark       & 0.92      & 1        & 1         & 0.9195    & 1        & 1         \\
		Thunderb. & 0.955     & 0.993    & 0.993     & 0.9835    & 1        & 1         \\
		Windows     & 0.997     & 1        & 1         & 0.759     & 1        & 1         \\
		Zookeeper   & 0.9665    & 1        & 1         & 0.972     & 1        & 1         \\ \hline
	\end{tabular}
\end{table}
	
\begin{table}[htb]
	\caption{Accuracy (Drain, sub-opt parameter)} 
	\label{Tb:Accuracy:DrainSubOpt} 
	\centering
		\begin{tabular}{p{1.1cm}cccccc}
			\hline
			\multirow{2}{*}{\textbf{Dataset}} & \multicolumn{3}{c}{\textbf{GA}}                    & \multicolumn{3}{c}{\textbf{MA}}                    \\ \cline{2-7} 
			& \textbf{drain} & \textbf{simu.} & \textbf{rpt} & \textbf{drain} & \textbf{simu.} & \textbf{rpt} \\
			\hline
			Andriod     & 0.712     & 0.998    & 0.998     & 0.7885    & 0.9995   & 0.9995    \\
			Apache      & 1         & 1        & 1         & 1         & 1        & 1         \\
			BGL         & 0.9115    & 1        & 1         & 0.918     & 1        & 1         \\
			Hadoop      & 0.962     & 1        & 1         & 0.963     & 1        & 1         \\
			HDFS        & 0.9975    & 1        & 1         & 1         & 1        & 1         \\
			HealthApp   & 0.78      & 1        & 1         & 0.9005    & 1        & 1         \\
			HPC         & 0.887     & 1        & 1         & 0.8965    & 1        & 1         \\
			Linux       & 0.681     & 0.8785   & 1         & 0.7425    & 0.941    & 0.941     \\
			Mac         & 0.6495    & 0.8995   & 1         & 0.7245    & 0.99     & 0.99      \\
			OpenSSH     & 0.718     & 1        & 1         & 0.717     & 1        & 1         \\
			OpenStack   & 0.2775    & 0.956    & 1         & 0.1955    & 1        & 1         \\
			Proxifier   & 0.0255    & 1        & 1         & 0.499     & 1        & 1         \\
			Spark       & 0.92      & 1        & 1         & 0.9195    & 1        & 1         \\
			Thunderb. & 0.947     & 0.993    & 0.993     & 0.973     & 1        & 1         \\
			Windows     & 0.568     & 1        & 1         & 0.4485    & 1        & 1         \\
			Zookeeper   & 0.9665    & 1        & 1         & 0.972     & 1        & 1         \\ \hline
		\end{tabular}
	\end{table}

\subsection{Evaluating Each Algorithmic Building Block}		
										
\noindent
{\bf Evaluating \texttt{Message-Completion}.} 
Table \ref{Tb:MC:Drain} shows the performance of our  \texttt{Message-Completion} algorithm, 
where we consider a weak version \texttt{Message-Completion} without seeking user feedback on 
dummy tokens.  
The column labeled ``base'' (or ``w/ mc'') denotes the 
fraction of message-complete pure clusters among all pure clusters 
under the base template mining algorithm such as Drain 
(or \texttt{Message-Completion}).  
When the base template mining algorithm is 
run with fine tuned parameters, 
 the \texttt{Message-Completion} algorithm makes all pure clusters 
have the message-complete template, 
no matter whether the fraction of pure clusters with the message-loss template 
is high or low.  
This statement also holds when base template mining algorithms 
are run with sub-optimal parameters.  
This shows \texttt{Message-Completion} algorithm 
is high accuracy in completing messages for pure clusters.  
It also validates Condition (\ref{eq:No-Loss-LCS}).  

\begin{table}[htb]
	\caption{Evaluating \texttt{Message-Completion}}
	\label{Tb:MC:Drain}
	\centering
	\begin{tabular}{ccccc}
		\hline
		\multirow{2}{*}{\textbf{Dataset}} & \multicolumn{2}{c}{\textbf{fine tune}}            & \multicolumn{2}{c}{\textbf{sub-opt}}         \\ \cline{2-5} 
		& \textbf{base} & \textbf{w/ mc} & \textbf{base} & \textbf{w/ mc} \\ \hline
		Andriod                           & 1                     & 1                    & 1                     & 1                    \\
		Apache                            & 1                     & 1                    & 1                     & 1                    \\
		BGL                               & 1                     & 1                    & 1                     & 1                    \\
		Hadoop                            & 1                     & 1                    & 1                     & 1                    \\
		HDFS                              & 1                     & 1                    & 1                     & 1                    \\
		HealthApp                         & 1                     & 1                    & 1                     & 1                    \\
		HPC                               & 1                     & 1                    & 1                     & 1                    \\
		Linux                             & 1                     & 1                    & 1                     & 1                    \\
		Mac                               & 0.9634                & 1                    & 0.9422                & 1                    \\
		OpenSSH                           & 0.9524                & 1                    & 0.9474                & 1                    \\
		OpenStack                         & 0.8267                & 1                    & 0.9                   & 1                    \\
		Proxifier                         & 1                     & 1                    & 1                     & 1                    \\
		Spark                             & 0.96                  & 1                    & 0.96                  & 1                    \\
		Thunderbird                       & 1                     & 1                    & 1                     & 1                    \\
		Windows                           & 0.7358                & 1                    & 0.7647                & 1                    \\
		Zookeeper                         & 1                     & 1                    & 1                     & 1                    \\ \hline
	\end{tabular}
\end{table}

\noindent
{\bf Evaluating \texttt{Merge}.} 
Section \ref{sec:exp:overall} shows drastic improvements in MA by Algorithm \ref{Algo:Combination}, 
which implies that the \texttt{Merge} algorithm has high accuracy.  
Thus, here we focus on understanding the query or human feedback complexity of the \texttt{Merge} algorithm.  
Table \ref{Tb:MergeFine} shows the query complexity of our \texttt{Merge} algorithm, 
where the base template mining algorithm Drain is run with 
fine tuned parameters.  
The column labeled ``\# of clst'' (or ``redun.'') corresponds 
to the total number of distinct pure clusters that have the different ground-truth template  
(or the redundancy of each distinct pure cluster, i.e., 
the average number of clusters have the same ground truth template).  
One can observe the number of distinct pure clusters varies 
from tens to hundreds.  
The redundancy varies from 1 to around 5.  
The column labeled ``\# of fb'' corresponds to the total 
number of human feedback in merging these pure clusters.  
Note that the user feedback is only needed for inaccurate templates.  
One can observe the total number of human feedback is 
slightly larger than the number of distinct pure templates 
regardless of the redundancy.  
The average (or maximum) number of additional feedback required by our method is 0.17 (or 0.4) times 
of the number of ground-truth templates.  
The statement also holds when the base template mining algorithm 
is run with sub-optimal parameters 
as shown in Table \ref{Tb:MergeSubOpt}. 
This further validates our theoretical findings that 
the query complexity of \texttt{Merge} algorithm 
is invariant of the total number of pure clusters.    
The column labeled ``avg. que. length'' (or ``avg. \# of cps'') corresponds to 
the average number of candidates in the question set 
(or the average number of comparisons that the user 
needs to select a similar template).  
One can observe that the average number of comparisons is around 1.  
It implies that when there exists a similar template, 
it is always sorted as top-1 in the question list.  
This saves the comparison costs of users.  
In our user study, participants take an average of 1.54s to answer each question, which is acceptable.  
Furthermore, the number of candidates in the question list 
is usually less than ten.  
Similar observations can be found 
when the base template mining algorithm 
is run with sub-optimal parameters 
as shown in Table \ref{Tb:MergeSubOpt}. 
These results show the low query complexity of the \texttt{Merge} algorithm.

\begin{table}[htb]
	\caption{Evaluating \texttt{Merge} (Drain, fine tune parameter)} 
	\label{Tb:MergeFine}
	\centering
	\begin{tabular}{cccccc}
		\hline
		\textbf{Dataset} & \textbf{\begin{tabular}[c]{@{}c@{}}\# of \\ clst\end{tabular}} & \textbf{\begin{tabular}[c]{@{}c@{}}  \\ redun.\end{tabular}} & \textbf{\begin{tabular}[c]{@{}c@{}}avg. que. \\ length\end{tabular}} & \textbf{\begin{tabular}[c]{@{}c@{}}avg. \#\\ of cps\end{tabular}} & \textbf{\begin{tabular}[c]{@{}c@{}}\# of \\ fb\end{tabular}} \\ \hline
		Andriod          & 146           & 1.0753     & 4.2635      & 1      & 149         \\
		Apache           & 6        & 1        & 2.4      & 0    & 6     \\
		BGL              & 96  & 1.0208  & 7.4227   & 1         & 98  \\
		Hadoop           & 91  & 1.022 & 6.3696  & 1 & 93    \\
		HDFS             & 14 & 1.1429  & 4.8462 & 0    & 14   \\
		HealthApp        & 69   & 4.4783   & 3.2319    & 1   & 70   \\
		HPC              & 37 & 1.1892  & 2.2051      & 1      & 40      \\
		Linux            & 106     & 1.0094     & 3.4057  & 1 & 107  \\
		Mac              & 314  & 1.2102 & 11.9036 & 1 & 333  \\
		OpenSSH          & 21 & 1  & 7.25    & 0 & 21   \\
		OpenStack        & 28  & 1.7143  & 5.1071   & 1 & 29     \\
		Proxifier        & 8    & 2.125  & 3.7143   & 0    & 8    \\
		Spark            & 25  & 1  & 2.0417   & 0  & 25  \\
		Thunderbird      & 132   & 1.3182  & 3.9779   & 1   & 137  \\
		Windows          & 50  & 1.06  & 4.1961  & 1  & 52  \\
		Zookeeper        & 42  & 1.0476 & 3.8293 & 0  & 42  \\ 
		\hline
	\end{tabular}
\end{table}
											
\begin{table}[htb]
	\caption{Evaluating \texttt{Merge} (Drain, sub-opt parameter)} 
	\label{Tb:MergeSubOpt}
	\centering
	\begin{tabular}{cccccc}
		\hline
		\textbf{Dataset} & \textbf{\begin{tabular}[c]{@{}c@{}}\# of \\ clst\end{tabular}} & \textbf{\begin{tabular}[c]{@{}c@{}} \\ redun.\end{tabular}} & \textbf{\begin{tabular}[c]{@{}c@{}}avg. que. \\ length\end{tabular}} & \textbf{\begin{tabular}[c]{@{}c@{}}avg. \# \\ of cps\end{tabular}} & \textbf{\begin{tabular}[c]{@{}c@{}}\# of \\ fb\end{tabular}} \\ \hline
		Andriod        & 151 & 1.0662  & 4.6     & 1  & 156   \\
		Apache         & 6   & 2.1667  & 2.1667  & 1  & 7     \\
		BGL            & 111 & 3.8919  & 8.1129  & 1  & 125   \\
		Hadoop         & 110 & 1.8727  & 8.375   & 1  & 129   \\
		HDFS           & 14  & 11.7857 & 3.9412  & 1  & 18    \\
		HealthApp      & 75  & 15.6267 & 3.1264  & 1  & 88    \\
		HPC            & 43  & 4.2093  & 2.0556  & 1  & 55    \\
		Linux          & 111 & 1.3784  & 3.2377  & 1  & 123   \\
		Mac            & 322 & 1.5062  & 11.2258 & 1  & 373   \\
		OpenSSH        & 23                                                                  & 22.3913                                                           & 6.8519                                                                       & 1                                                                              & 28                                                                   \\
		OpenStack        & 33                                                                  & 1                                                                 & 6.1562                                                                       & 0                                                                              & 33                                                                   \\
		Proxifier        & 8                                                                   & 83.375                                                            & 3.375                                                                        & 1                                                                              & 9                                                                    \\
		Spark            & 29                                                                  & 35.069                                                            & 2.1282                                                                       & 1                                                                              & 40                                                                   \\
		Thunderbird      & 148                                                                 & 1.7027                                                            & 4.0602                                                                       & 1                                                                              & 167                                                                  \\
		Windows          & 50                                                                  & 1.32                                                              & 3.9636                                                                       & 1                                                                              & 56                                                                   \\
		Zookeeper        & 46                                                                  & 1.3478                                                            & 3.8113                                                                       & 1                                                                              & 54                                                                   \\ \hline
	\end{tabular}
\end{table}

\noindent
{\bf Evaluating \texttt{Separation}.}  
We focus on understanding the query or human feedback complexity of the \texttt{Separation} algorithm.  
Table \ref{Tb:Separation:Fine} shows the query complexity of our \texttt{Separation} algorithm, 
where the base template mining algorithm Drain is run with 
fine tuned parameters.  
The column labeled ``\# of tpl'' (or ``redun.'') corresponds 
to the total number of distinct ground-truth templates   
(or the redundancy of each template, i.e., 
the average number of logs associated with each template).  
One can observe the number of distinct templates varies 
from zero to around twenty.  
The redundancy varies from 0 to around twenty.  
The column labeled ``\# of fb'' corresponds to the average 
number of human feedback in separating each cluster of logs.  
One can observe the total number of human feedback is 
less than four regardless of the redundancy.  
This statement also holds when the base template mining algorithm 
is run with sub-optimal parameters, 
as shown in Table \ref{Tb:Separation:Sub}. 
This further validates our theoretical findings that 
the query complexity of the \texttt{Separation} algorithm 
is invariant of the total number of logs.    
The column labeled `avg. que. length'' (or ``avg. \# of cps'') corresponds to 
the average number of candidates in the question set 
(or the average number of comparisons that the user 
needs to select a similar template).  
One can observe that the average number of comparisons is around 1.  
This implies that when there exists a similar template,  
it is always sorted as top-1 in the question list.  
This saves the comparison costs of users.  
Furthermore, the number of candidates in the question list 
is usually less than three.  
Similar observations can be found 
when the base template mining algorithm 
is run with sub-optimal parameters, 
as shown in Table \ref{Tb:Separation:Sub}. 
These results show the low query complexity of the \texttt{Separation} algorithm.

\begin{table}[htb]
	\caption{Evaluating \texttt{Separation} (Drain,Fine tune parameter)}
	\label{Tb:Separation:Fine}
	\centering
	\begin{tabular}{cccccc}
		\hline
		\textbf{Dataset} & \textbf{\begin{tabular}[c]{@{}c@{}} \# of \\ tpl \end{tabular}} & \textbf{\begin{tabular}[c]{@{}c@{}} \\ redun.\end{tabular}} & \textbf{\begin{tabular}[c]{@{}c@{}}avg. que. \\ length\end{tabular}} & \textbf{\begin{tabular}[c]{@{}c@{}}avg. \# \\ of cps\end{tabular}} & \textbf{\begin{tabular}[c]{@{}c@{}}avg. \#\\of fb\end{tabular}} \\ \hline
		Andriod          & 20                                                               & 1.5                                                          & 1.5294                                                                   & 1                                                                              & 1.5455                                                         \\
		Apache           & 0                                                                & 0                                                            & 0                                                                        & 0                                                                              & 0                                                              \\
		BGL              & 23                                                               & 1.5217                                                       & 2.2222                                                                   & 1                                                                              & 2                                                              \\
		Hadoop           & 22                                                               & 3.4091                                                       & 2.3636                                                                   & 1                                                                              & 2.75                                                           \\
		HDFS             & 0                                                                & 0                                                            & 0                                                                        & 0                                                                              & 0                                                              \\
		HealthApp        & 6                                                                & 1                                                            & 2                                                                        & 0                                                                              & 1                                                              \\
		HPC              & 11                                                               & 10.4545                                                      & 1.4375                                                                   & 1                                                                              & 3.2                                                            \\
		Linux            & 10                                                               & 6.1                                                          & 1.4167                                                                   & 1                                                                              & 2                                                              \\
		Mac              & 25                                                               & 2.56                                                         & 1.65                                                                     & 1                                                                              & 1.5385                                                         \\
		OpenSSH          & 6                                                                & 3.1667                                                       & 1.5                                                                      & 1                                                                              & 2                                                              \\
		OpenStack        & 17                                                               & 21.2941                                                      & 1.5185                                                                   & 1                                                                              & 3.8571                                                         \\
		Proxifier        & 0                                                                & 0                                                            & 0                                                                        & 0                                                                              & 0                                                              \\
		Spark            & 11                                                               & 14.3636                                                      & 2.1818                                                                   & 1                                                                              & 2.75                                                           \\
		Thunderbird      & 17                                                               & 1.3529                                                       & 2.0909                                                                   & 1                                                                              & 1.375                                                          \\
		Windows          & 0                                                                & 0                                                            & 0                                                                        & 0                                                                              & 0                                                              \\
		Zookeeper        & 8                                                                & 6.75                                                         & 2.4545                                                                   & 1                                                                              & 5.5                                                            \\ \hline
	\end{tabular}
\end{table}

\begin{table}[htb]
	\caption{Evaluating \texttt{Separation} (Drain,Sub-Opt parameter)}
	\label{Tb:Separation:Sub}
	\centering
	\begin{tabular}{cccccc}
		\hline
		\textbf{Dataset} & \textbf{\begin{tabular}[c]{@{}c@{}}\# of \\ tpl\end{tabular}} & \textbf{\begin{tabular}[c]{@{}c@{}}\\ redun.\end{tabular}} & \textbf{\begin{tabular}[c]{@{}c@{}}avg. que. \\ length\end{tabular}} & \textbf{\begin{tabular}[c]{@{}c@{}}avg. hit \\ length\end{tabular}} & \textbf{\begin{tabular}[c]{@{}c@{}}avg. \#\\ of fb\end{tabular}} \\ \hline
		Andriod          & 42                                                               & 2.2143                                                       & 1.6667                                                                   & 1                                                                              & 1.5714                                                         \\
		Apache           & 0                                                                & 0                                                            & 0                                                                        & 0                                                                              & 0                                                              \\
		BGL              & 38                                                               & 2.0263                                                       & 2.069                                                                    & 1                                                                              & 1.8125                                                         \\
		Hadoop           & 22                                                               & 3.4091                                                       & 3.5833                                                                   & 1                                                                              & 4                                                              \\
		HDFS             & 0                                                                & 0                                                            & 0                                                                        & 0                                                                              & 0                                                              \\
		HealthApp        & 6                                                                & 1                                                            & 2                                                                        & 0                                                                              & 1                                                              \\
		HPC              & 11                                                               & 10.4545                                                      & 1.4375                                                                   & 1                                                                              & 3.2                                                            \\
		Linux            & 18                                                               & 3.8333                                                       & 1.5625                                                                   & 1                                                                              & 1.6                                                            \\
		Mac              & 90                                                               & 3.8111                                                       & 1.7692                                                                   & 1                                                                              & 2.3333                                                         \\
		OpenSSH          & 8                                                                & 16.5                                                         & 1.4444                                                                   & 1                                                                              & 2.25                                                           \\
		OpenStack        & 25                                                               & 56.88                                                        & 1.95                                                                     & 1                                                                              & 5.7143                                                         \\
		Proxifier        & 4                                                                & 60                                                           & 1.3333                                                                   & 1                                                                              & 3                                                              \\
		Spark            & 11                                                               & 14.3636                                                      & 2.1818                                                                   & 1                                                                              & 2.75                                                           \\
		Thunderbird      & 27                                                               & 1.5556                                                       & 2.1111                                                                   & 1                                                                              & 1.5                                                            \\
		Windows          & 14                                                               & 44.5                                                         & 1.7778                                                                   & 1                                                                              & 1.2857                                                         \\
		Zookeeper        & 8                                                                & 6.75                                                         & 2.4545                                                                   & 1                                                                              & 5.5                                                            \\ \hline
	\end{tabular}
\end{table}
								
\subsection{Evaluating the Running Time}

We evaluate the running time of our proposed algorithms 
on the BGL dataset \cite{zhu2019tools}, 
which contains over 4 million lines of logs.  
The BGL dataset was widely used in previous works \cite{chu2021prefix,zhu2019tools} 
to evaluate the running time of log parsing algorithms.  
We run each base algorithm using its fine tuned parameters \cite{zhu2019tools, chu2021prefix}.   
We run a sequential combination of \texttt{Message-Completion}, \texttt{Merge} and \texttt{Separation}   
on the output of each base algorithm.  
The total running time of this combination serves as the running time of our proposed method.    
Table \ref{Tb:Efficiency} shows the running time of base algorithms and our proposed method, 
where we vary the lines of logs from $10^4$ (the first $10^4$ lines of logs of BGL) to $10^6$ (the first $10^6$ lines of  logs of BGL).  
Considering the Drain algorithm, 
one can observe that under $10^4$/$10^5$/$10^6$ lines of logs, 
our method's total running time is  3.4s/14.3s/35.6s.    
Furthermore, under $10^4$/$10^5$/$10^6$ lines of logs, 
our method's total running time is 70\%/44\%/25\% of that of the Drain algorithm.  
This shows that as the lines of logs increase, 
the computational efficiency of our method increases compared to the base algorithm.  
Similar running time comparisons can also be 
observed on the other four base algorithms.  
These results show that our method is computationally efficient.   

\begin{table}[htb]
\caption{Evaluating the Running Time}
\label{Tb:Efficiency}
\centering
\small
\begin{tabular}{cccccc}
\hline
\multirow{2}{*}{\#logs} & \multicolumn{5}{c}{Algorithm(Base/Ours)}                        \\ \cline{2-6} 
                          & Drain      & Spell      & IPLoM      & Logram     & PrefixGraph \\ \hline
$10^4$                     & 4.8/3.4    & 3.8/2.4    & 4.8/2.7    & 5.2/3.5    & 3.4/2.5     \\
$10^5$                    & 32.2/14.3  & 31.4/13.3  & 31.6/14.9  & 42.7/21.4  & 35.3/13.9   \\
$10^6$                   & $\!\!$145.9/35.6 & $\!\!$162.3/32.2 & $\!\!$149.3/36.3 & $\!\!$158.4/52.4 & $\!\!$154.9/34.8  \\ \hline
\end{tabular}
\end{table}

\section{conclusion}

This paper develops a human-in-the-loop template mining framework to support 
interactive log analysis.  
We formulated three types of light-weight user feedback, 
and based on them we designed three atomic human-in-the-loop template mining algorithms.  
We derived mild conditions under which the output of 
our proposed algorithms are provably correct.  
We also derived upper bounds on the computational complexity and 
query complexity of each algorithm.  
Extensive experiments demonstrated the versatility and 
efficiency of our proposed algorithms.

	
\begin{acks} 
This work was supported in part by Alibaba Innovative Research grant
(ATA50DHZ4210003), the RGC’s GRF (14200321), 
Chongqing Talents: Exceptional Young Talents Project (cstc2021ycjhbgzxm0195).  
Hong Xie is the corresponding author.  
\end{acks}
\bibliographystyle{ACM-Reference-Format}
\bibliography{references}

\appendix
\section*{Appendix}

\section{Technical Proofs}

\noindent
{\bf Proof of Theorem \ref{thm:MC:correctness}: } 
Note that Condition (\ref{eq:No-Loss-LCS}) indicates that 
longest subsequence of 
two logs that have the same template summarizes extracts complete message of two these two logs.  
Thus, if $\widehat{\mathcal{C}}_k$ is pure, the final output satisfies that $\bm{T}_{\text{mc}} \in \mathcal{S}_{\text{complete}}$.  
Otherwise, $\bm{T}_{\text{mc}}$ is partial, i.e., $\bm{T}_{\text{mc}} \in \mathcal{S}_{\text{partial}}$
\done

\noindent
{\bf Proof of Lemma \ref{lem:LosslessTemplate}: }  
Note that each round, at least one token is removed from both $\bm{a}$ and $\bm{b}$.  
Thus, Algorithm \ref{Algo:Human-CSS} terminates in at most $\min\{ \texttt{len}(\bm{a}), \texttt{len}(\bm{a}) \}$ rounds.  
If user does not make errors in providing feedbacks, i.e., 
they identify dummy tokens, 
all dummy tokens will be removed.  
This implies that $\bm{T} \sqsubseteq \widehat{\bm{T}}$. 
\done

\noindent
{\bf Proof of Theorem \ref{thm:MC:ComComplex}:} 
It is a simple consequence that 
the computational complexity of matching is $O(\widehat{M}_{\text{max}})$
and the computational complexity of 
extracting longest common subsequence is 
$O(\widehat{M}^2_{\text{max}})$
\done

\noindent
{\bf Proof of Theorem \ref{thm:merge:correctness}: }
Eq. (\ref{eq:Merge:AllHandle}) is a consequence that each template is processed 
and it is either placed in message-loss set of message-complete set.  
Eq. (\ref{eq:Merge:loss}) is a consequence that users provide correct 
feedbacks on whether a template has full message of has message loss.  
Eq. (\ref{eq:Merge:Completeloss})-(\ref{eq:Merge:CompleteRedundancy}) are a 
consequence that users provide correct feedbacks on templates with the same messages 
and templates with the same messages are combined without message loss.  
\done

\noindent
{\bf Proof of Theorem \ref{thm:Merge:computational}: } 
We only analyze the computational complexity of the following dominant parts.   

For each template, step \ref{step:merge:match} searches a matching template for it.  
The size of the searching set $\mathcal{P}_{\text{complete}} $ 
is at most $N^{\text{dst}}_{\text{mg}}$.  
The complexity of matching is $O(\widetilde{M}_{\text{max}})$.  
Thus, the total computational complexity for searching is 
upper bounded by 
\[
|\mathcal{P}_{\text{mg}}| 
N^{\text{dst}}_{\text{mg}}
\widetilde{M}_{\text{max}}.  
\]

For each message-complete template, 
step \ref{step:merge:select} requests the user to 
select a template that has the same message with it.  
As shown in Theorem \ref{thm:Merge:Query}, 
the total number of human-select queries is upper bounded by 
$N^{\text{dst}}_{\text{mg}}  ( \widetilde{d}_{\text{max}} +1)$
This selection process involves sorting templates in 
$\mathcal{T}_{\text{complete}}$, 
which contains at most $N^{\text{dst}}_{\text{mg}} $ templates.  
Thus the computational complexity of sorting is $N^{\text{dst}}_{\text{mg}}  \ln N^{\text{dst}}_{\text{mg}}$.  
Furthermore, the sorting is based on the longest common subsequence whose 
computational complexity is upper bounded by 
$
N^{\text{dst}}_{\text{mg}}  \widetilde{M}^2_{\text{max}}
$
The total computational complexity of human-select queries is upper bounded by 
\begin{align*}
& 
N^{\text{dst}}_{\text{mg}}  ( \widetilde{d}_{\text{max}} +1)
\big(
N^{\text{dst}}_{\text{mg}}  \ln N^{\text{dst}}_{\text{mg}}
+
N^{\text{dst}}_{\text{mg}}  \widetilde{M}^2_{\text{max}}
\big)
\\
& 
= ( \widetilde{d}_{\text{max}} +1) 
\left(
N^{\text{dst}}_{\text{mg}}
\right)^2 
\big(\ln N^{\text{dst}}_{\text{mg}}
+
\widetilde{M}^2_{\text{max}}
\big).  
\end{align*}

Step \ref{step:merge:merge} utilizes human-dummy-token feedbacks 
to extract a new template.  
Extracting a human-dummy-token feedback 
involves computing a longest common subsequence between two templates, 
whose computational complexity is upper bounded by $\widetilde{M}^2_{\text{max}}$.  
As shown in Theorem \ref{thm:Merge:Query}, 
the total number of human-dummy-token queries is upper bounded by 
$
N^{\text{dst}}_{\text{mg}}  
\widetilde{d}^2_{\text{max}}
$
Thus the total computational complexity of human-dummy-token queries is 
upper bounded by
\[
N^{\text{dst}}_{\text{mg}}  
\widetilde{d}^2_{\text{max}}
\widetilde{M}^2_{\text{max}}.  
\]
This proof is then complete.   
\done

\noindent
{\bf Proof of Theorem \ref{thm:Merge:Query}: }   
For each template from the input set $\mathcal{P}_{\text{mg}}$, 
step \ref{step:merge:loss} one user feedback on whether it 
has message loss or not.  
Thus, the total number of human-message-loss query is 
\[
\# \text{ of } \texttt{Human-Message-Loss} \text{ feedback }
= 
|\mathcal{P}_{mg}|.  
\] 

For each template with complete message, 
step \ref{step:merge:select} request the user to select 
a template from the set $\mathcal{T}_{\text{complete}}$ 
that has the same message as it.    
Note that each human-select feedback shortens the 
candidate template in the set $\mathcal{T}_{\text{complete}}$ by at least one token, 
except in the initializing of a new template in step \ref{step:merge:nonNewClus}.  
In the initializing of a new template, one human-select query is taken.    
Thus, the total number of human-select query is upper bounded by 
\[
\# \text{ of } \texttt{Human-Select} \text{ feedback } 
= 
N^{\text{dst}}_{\text{mg}}  ( \widetilde{d}_{\text{max}} +1).  
\]
  
Step \ref{step:merge:merge} utilizes human-dummy-token feedbacks 
to extract a new template.  
Each extraction of new template needs at most $\widetilde{d}_{\text{max}}$
human-dummy-token feedbacks.  
Note that each extraction of new template shortens the 
candidate template in the set $\mathcal{T}_{\text{complete}}$ by at least one token.  
The total number extracting new template is upper bounded 
by $N^{\text{dst}}_{\text{mg}}  
\widetilde{d}_{\text{max}}$.  
Thus, the total number of human-dummy-token queries is upper bounded by: 
\[
\# \text{ of } \texttt{Human-Dummy-Token} \text{ feedback }  
\leq 
N^{\text{dst}}_{\text{mg}}  
\widetilde{d}_{\text{max}} 
\times 
\widetilde{d}_{\text{max}} 
= N^{\text{dst}}_{\text{mg}}  
\widetilde{d}^2_{\text{max}}.  
\] 
This proof is then complete. 
\done

\noindent
{\bf Proof of Theorem \ref{thm:separation:correct}: }  
The proof of this theorem is similar to that of   Theorem \ref{thm:merge:correctness}.  
\done

\noindent
{\bf Proof of Theorem \ref{thm:Separation:QueryComp}: } 
Note that each human-select query shorten the 
the candidate template by at least one token, 
except in the step of initializing a candidate template.  
The candidate template is initialized as a log as shown in  
step \ref{Step:Separation:InitialTemplate} of Algorithm \ref{Algo:Separation}.  
There are at most $N^{\text{tpl}}_{\text{sep}} $ distinct candidate templates.  
Thus, the number of human-select query is upper bounded by 
\[
\# \text{ of } \texttt{Human-Select} \text{ feedback } 
\leq 
N^{\text{tpl}}_{\text{sep}}   
(d_{\text{max}} +1).  
\]
 
Step \ref{Step:Separation:LosslessTemplate} 
of Algorithm \ref{Algo:Separation} involves human-dummy-token queries.  
In particular, each such query delete at least one tokens.  
Thus, one call of Step \ref{Step:Separation:LosslessTemplate} 
incurs at most $d_{\text{max}} $ human-dummy-token queries.  
Each calling of Step \ref{Step:Separation:LosslessTemplate} 
shorten the candidate template by at least one token.  
Thus, the total number of calling Step \ref{Step:Separation:LosslessTemplate} 
is upper bounded by 
$N^{\text{tpl}}_{\text{sep}} d_{\text{max}}$.  
Hence, the the total number of human-dummy-token queries is upper bounded by 
\begin{align*}
\# \text{ of } \texttt{Human-Dummy-Token} \text{ feedback }
& 
\leq 
N^{\text{tpl}}_{\text{sep}} d_{\text{max}}  \times d_{\text{max}} 
\\
& 
= 
N^{\text{tpl}}_{\text{sep}} d^2_{\text{max}}.
\end{align*}
This proof is then complete.  
\done

\noindent
{\bf Proof of Theorem \ref{thm:Separation:Comp}: }
We only analyze the computational complexity of the following dominant parts.   

For each log, step \ref{Step:Separation:Matching} 
conducts a matching to each template in $\mathcal{P}_{\text{sep} }$.  
The set $\mathcal{P}_{\text{sep} }$ has at most 
$N^{\text{tpl}}_{\text{sep}}$ templates.   
The computational complexity of conduction one matching is $O(M_{\text{max}})$. 
Thus the total computational complexity of conduct matching operations is 
upper bounded by
\[
\left| 
\widehat{\mathcal{C}}_{\text{sep}}
\right|  
N^{\text{tpl}}_{\text{sep}}
M_{\text{max}}.  
\] 

Step \ref{Step:Separation:Select} conducts selection queries.  
As shown in Theorem \ref{thm:Separation:QueryComp}, 
the number of human-select queries is upper bounded by 
$
N^{\text{tpl}}_{\text{sep}}  (d_{\text{max}} +1).  
$
The computational of each human-select query involves 
sorting templates in $\mathcal{T}_{\text{sep}}$, 
which contains at most $N^{\text{tpl}}_{\text{sep}}$ templates.  
Thus the sorting complexity is upper bounded by 
$
N^{\text{tpl}}_{\text{sep}} 
\ln 
N^{\text{tpl}}_{\text{sep}}
$.    
Furthermore, the sorting is based on the longest common subsequence 
whose computation complexity is upper bounded by 
\[
N^{\text{tpl}}_{\text{sep}} M^2_{\text{max}}
\] 
The total computational complexity of human-select queries is upper bounded by 
\begin{align*}
& 
N^{\text{tpl}}_{\text{sep}}  (d_{\text{max}} +1) 
\left(
 N^{\text{tpl}}_{\text{sep}} 
\ln 
N^{\text{tpl}}_{\text{sep}}
+ 
N^{\text{tpl}}_{\text{sep}} M^2_{\text{max}}
\right)
\\
& 
=
(d_{\text{max}} +1) 
\left( N^{\text{tpl}}_{\text{sep}} \right)^2
\left( 
\ln 
N^{\text{tpl}}_{\text{sep}}
+ 
M^2_{\text{max}}
\right).  
\end{align*}

Step \ref{Step:Separation:LosslessTemplate} involves human-dummy-token queries.  
As shown in Theorem \ref{thm:Separation:QueryComp}, 
the number of human-dummy-token queries is upper bounded by 
$
N^{\text{tpl}}_{\text{sep}} d^2_{\text{max}}.  
$
Each dummy involves one longest common subsequence operations.  
Thus, the total computational complexity of 
human-dummy-token queries is upper bounded by 
\[
N^{\text{tpl}}_{\text{sep}} d^2_{\text{max}} 
M^2_{\text{max}}. 
\]
This proof is then complete.  
\done 
\end{document}